\documentclass{adobe_research}
\usepackage{lmodern}
\usepackage{array}
\usepackage{colortbl}
\usepackage{amsmath,amssymb}
\usepackage{url}
\usepackage{enumitem}
\usepackage{float}
\usepackage{pgfplots}\pgfplotsset{compat=1.18}\usepgfplotslibrary{groupplots}\usetikzlibrary{calc}
\usepackage{wrapfig}
\usepackage{algorithm}
\usepackage{algpseudocode}
\hypersetup{colorlinks=true,linkcolor=blue,citecolor=adobered,urlcolor=blue}

\newcommand{\method}{DMA\textsuperscript{2}}
\definecolor{methbg}{HTML}{FCE3C9}
\newcommand{\qmh}{\begin{tabular*}{\linewidth}{@{}*{4}{>{\centering\arraybackslash}p{0.243\linewidth}}@{}}
\rowcolor{methbg}\scriptsize Teacher & \scriptsize DMD2 & \scriptsize Decoupled & \scriptsize \mbox{\method{} (ours)}\end{tabular*}}

\newcommand{\qpl}[1]{{\scriptsize\itshape ``#1''}}
\newcommand{\qps}[1]{{\footnotesize\itshape ``#1''}}
\newcommand{\qi}[1]{\includegraphics[width=\linewidth]{#1}}

\newcommand{\dpg}{DPG-Bench}
\newcommand{\geneval}{GenEval}

\title{DMA\textsuperscript{2}: Pixel-space Distribution Matching with Adversarial and Anchor Losses}

\author[1,2,*]{Xin Lin}
\author[2]{Zhifei Zhang}
\author[2]{Yuqian Zhou}
\author[2]{Haitian Zheng}
\author[2]{Shaoteng Liu}
\authorbreak
\author[2,3,*]{Lehan Yang}
\author[2]{Zhe Lin}
\author[4]{Ming-Hsuan Yang}
\author[1]{Truong Nguyen}

\affiliation[1]{UC San Diego}
\affiliation[2]{Adobe Research}
\affiliation[3]{University of Virginia}
\affiliation[4]{UC Merced}
\contribution[*]{Work done during an internship at Adobe Research}

\begin{document}
\raggedbottom
\newcommand{\renderteaser}{%
\begin{figure}[H]
\centering
\includegraphics[width=\linewidth]{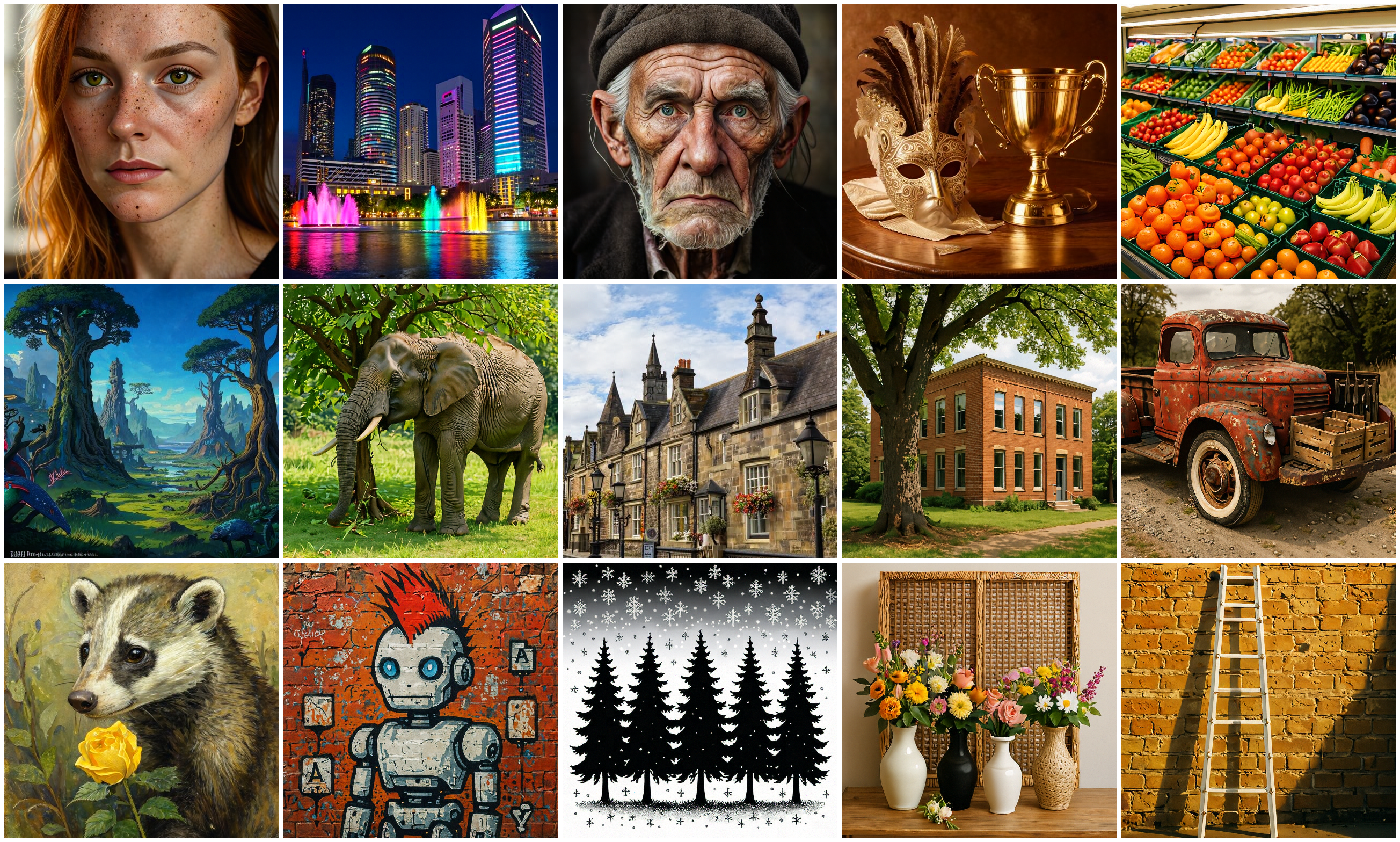}
\caption{Qualitative results of four-step text-to-image generation of \method{} ($512\times512$).}
\label{fig:teaser}
\end{figure}
\vspace{-0.6em}
}

\begin{adobeabstract}
Distribution matching distillation (DMD) provides a general framework for few-step diffusion generation, but its modern text-to-image instantiations have been developed primarily around latent diffusion.
It therefore overlooks key properties and design opportunities of native RGB. We revisit two DMD interfaces for pixel-space teachers. On the teacher-matching side, diagnostics show low-noise RGB matching is dominated by a local-texture cue, motivating a fixed high-noise matching band. On the real-data side, native clean-RGB outputs allow guidance from an external visual representation without traversing a decoder or sharing the heavy fake-score critic. DINO-Adv removes this critic from the adversarial gradient path and supplies local parametric patch guidance. For distribution-level guidance, we introduce AF-Loss, a parameter-free auxiliary semantic distribution-field objective designed for text-to-image DMD. It operates on detached rolling real and generated supports in the shared DINOv2 space while preserving prompt-conditioned teacher supervision. AF-Loss adds no learnable parameters or inference-time computation. Together these designs form \method{}.
Across DPG-Bench, GenEval, VQAScore, and COCO30K, the four-step \method{} student performs better than the 25-step teacher and evaluated few-step distillers.
\end{adobeabstract}

\renderteaser

\section{Introduction}\label{sec:intro}
Recently, pixel diffusion models~\citep{yu2025pixeldit, ma2025deco} advance text-to-image generation by modeling images end-to-end in RGB, avoiding the reconstruction ceiling and decoding artifacts of a VAE.
However, recent pixel diffusion still requires many sequential denoiser evaluations, making generation slow and costly.
Few-step distillation, represented by distribution matching distillation (DMD)~\citep{yin2023dmd, yin2024dmd2}, is the standard remedy: it uses a learned fake score to optimize a distribution-matching objective toward teacher outputs, instead of regressing the sampling trajectory step by step, which compounds error as steps are removed~\citep{luo2023lcm,xu2023instaflow}.
DMD is general, but its modern text-to-image recipe has been developed primarily around latent diffusion. Direct RGB generation changes both supervision interfaces: fine texture is explicit during teacher matching, while clean outputs expose pretrained visual representations for real-data guidance.

We analyze how the general DMD recipe interacts with native RGB. For teacher matching, the standard recipe samples matching levels broadly, without accounting for how information is represented in uncompressed RGB. Low-noise inputs already determine coarse image content, so teacher--critic differences can be dominated by fine texture rather than the distributional factors DMD should transfer. Our diagnostics indicate that higher-noise matching is more consistent with structural supervision.
For real-data guidance, DMD2 adds a GAN loss against real images~\citep{yin2024dmd2}. Its conventional Feature-GAN design attaches the adversarial head to the $560$M fake-score network. For a native-RGB student, this coupling is avoidable: clean outputs allow guidance through a pretrained visual representation without traversing a decoder or sharing the fake-score critic.

Motivated by this evidence, we design \method{}, which specializes the teacher-matching and real-data-guidance interfaces of DMD for native-RGB generation.
On the teacher-matching side, frequency controls, gradient visualizations, and a targeted matching-band intervention provide converging evidence for a near-clean local-texture bias in native-RGB matching (Section~\ref{sec:dmd}). These diagnostics motivate a fixed high-noise matching band that improves the model's performance without additional trainable modules or online scheduling.
On the real-data side, we exploit direct access to clean RGB through a dual-granularity, critic-decoupled guidance design. DINO-Adv removes the fake-score critic from the adversarial gradient path and uses an $86$M-parameter frozen DINOv2~\citep{oquab2023dinov2} encoder with only $1.58$M learnable parameters in its heads, reducing same-pipeline step time from $1.82$ to $0.72$ seconds ($2.5\times$ faster). This local adversary, however, does not define how each generated sample should move relative to the empirical real and generated distributions in semantic feature space. We therefore introduce Anchor-Field Loss (AF-Loss), a parameter-free auxiliary semantic distribution-field objective designed for text-to-image DMD. AF-Loss acts alongside the prompt-conditioned teacher objective, maintains detached rolling supports across minibatches, and shares the frozen DINOv2 encoder with DINO-Adv. It complements local adversarial guidance without adding learnable parameters or inference-time computation. With both sides addressed, the four-step student performs better than the 25-step DeCo \citep{ma2025deco} teacher and existing few-step distillers on all reported metrics; the same recipe transfers to a second backbone (PixelGen-XXL; Appendix~\ref{app:pixelgen}).
The main contributions of this work are:
\begin{itemize}[leftmargin=1.25em,itemsep=1pt,topsep=2pt]
    \item We propose \method{}, a few-step native-RGB DMD framework that specializes both teacher matching and real-data guidance for pixel-space generation. Across DPG-Bench, GenEval, VQAScore, and COCO30K, its four-step student ranks first among the evaluated few-step distillers and performs better than the $25$-step teacher on all seven reported metrics; the same recipe transfers to a second pixel-space backbone.
    \item We specialize both sides of the general DMD recipe for native RGB. On the teacher-matching side, converging diagnostics provide evidence for a near-clean local-texture shortcut and motivate \textbf{DM-Band}, which improves both alignment and image quality. On the real-data side, direct RGB access enables \textbf{DINO-Adv} to decouple local adversarial guidance from the $560$M learnable fake-score critic, using an $86$M-parameter frozen DINOv2 backbone and only $1.58$M learnable head parameters, while reducing same-pipeline step time from $1.82$ to $0.72$ seconds.
    \item We introduce \textbf{AF-Loss}, a parameter-free auxiliary semantic distribution-field objective designed for text-to-image DMD. It supplies distribution-level guidance from detached rolling supports while preserving prompt-conditioned teacher supervision, shares the frozen DINOv2 encoder with DINO-Adv, and adds no learnable parameters or inference-time computation.
\end{itemize}

\begin{figure}[t]
    \centering
    \includegraphics[width=\linewidth]{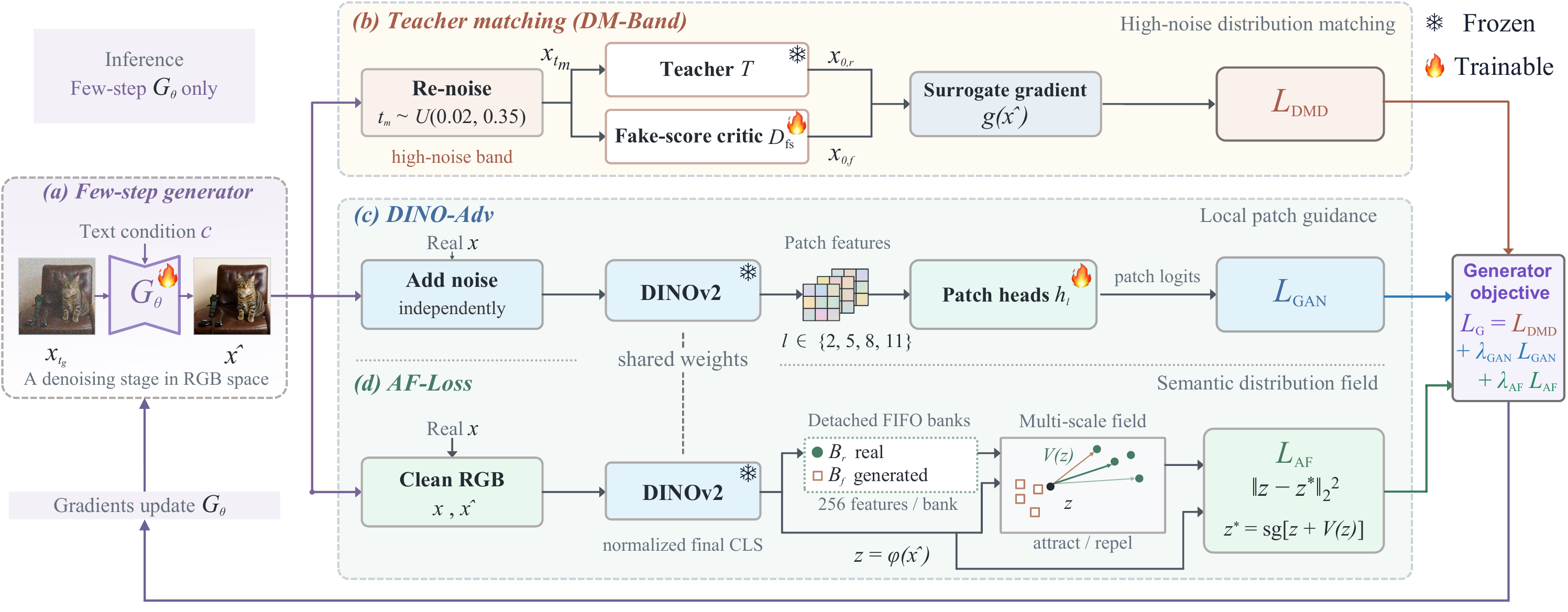}
    \caption{Overview of the proposed \method{} framework. DM-Band restricts teacher matching to a diagnostically selected high-noise range. For real-data guidance, DINO-Adv and AF-Loss share a frozen DINOv2 encoder and provide local patch guidance and a parameter-free semantic distribution-field target, respectively. Only $G_\theta$, the fake-score critic, and the adversarial heads are trained; all auxiliary components are removed at inference.}
    \label{fig:method}
\end{figure}

\section{Related Work}
\paragraph{Pixel-space diffusion.} A line of work generates in RGB without a frozen VAE, from cascaded pixel models~\citep{saharia2022imagen} to end-to-end RGB transformers competitive with latent models~\citep{jabri2023rin,hoogeboom2023simple,hoogeboom2025simpler,crowson2024hdit}. PixelDiT scales dual-level pixel transformers to megapixel generation~\citep{yu2025pixeldit}, whereas PiD uses conditional pixel diffusion for latent decoding and upsampling~\citep{lu2026pid}. The paradigm also extends to controllable generation, restoration, depth and geometry, and 3D~\citep{lin2026pixelcontrol,sun2026pixrestore,xu2025pixelperfectdepth,liu2026lapis,yuan2026pxdepth,xu2026pointdit,gao2026pixworld}. Our teacher DeCo~\citep{ma2025deco} is a frequency-decoupled member of this family. This literature primarily studies the training of native-RGB text-to-image models; few-step distillation tailored to their representation characteristics remains comparatively underexplored.

\paragraph{DMD-based diffusion distillation.} Distribution matching distillation trains a few-step student to match a teacher's output distribution through a learned fake score~\citep{yin2023dmd}, and DMD2 adds a real-image GAN term with two-time-scale updates for stability~\citep{yin2024dmd2}. Recent baselines Flash, TDM, Decoupled DMD, and DMDR span general acceleration, trajectory matching, CFG--DMD decoupling, and RL post-training, respectively~\citep{chadebec2025flash,luo2025tdm,decoupleddmd2026,jiang2026dmdr}. Related score and variational objectives distill in one step~\citep{zhou2024sid,nguyen2024swiftbrush}, while consistency and rectified-flow methods regress the sampling trajectory~\citep{luo2023lcm,xu2023instaflow}. Modern text-to-image distillation has centered on latent teachers; we specialize DMD's matching noise and real-data guidance for native RGB.

\paragraph{Feature-space objectives and field losses.} Frozen encoders support projected/DINOv2 discriminators~\citep{sauer2021projected,oquab2023dinov2}, representation alignment~\citep{yu2025repa}, and distributional objectives based on feature statistics or attract--repel fields~\citep{yang2026rfl,feng2026rdm,deng2026drifting}. Drifting Models use a kernelized field as the standalone objective for class-conditional generation~\citep{deng2026drifting}. AF-Loss is designed as an auxiliary objective for text-to-image DMD: prompt-conditioned teacher supervision remains active, detached rolling supports extend beyond the current minibatch, and the DINOv2 encoder is shared with DINO-Adv.

\section{Method}

\begin{figure}[t]
    \centering
    \includegraphics[width=\linewidth]{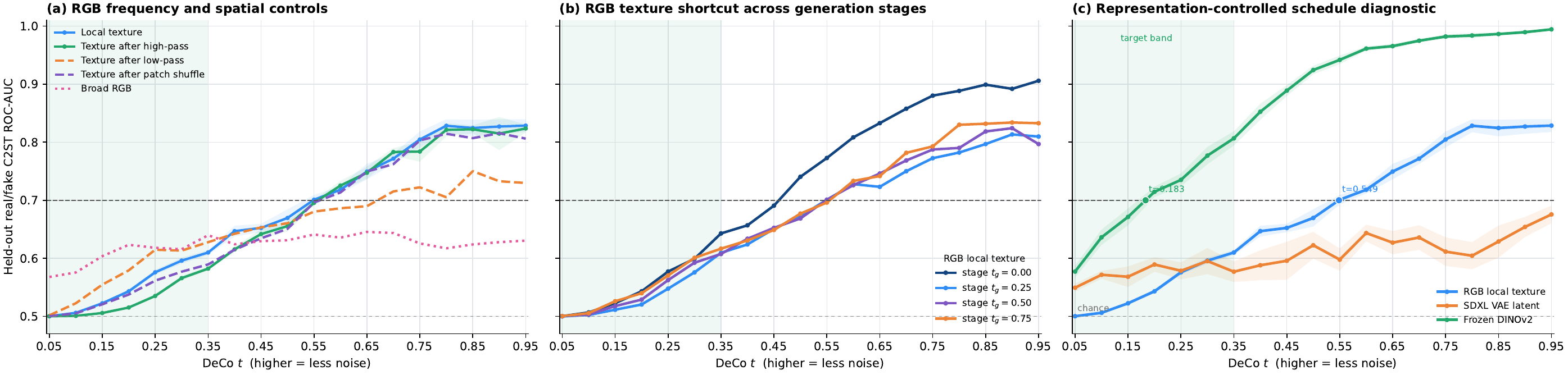}
    \caption{\textbf{Separability diagnostic (C2ST).} Each panel plots a held-out classifier's real/fake separability ($y$-axis: ROC-AUC over $1{,}536$ matched real/generated pairs, $0.5$ is chance) against the DeCo matching level $t$ ($x$-axis; larger $t$ is less noise). The panels vary the setup: \emph{(a)} the RGB input transform (raw texture, high-pass, low-pass, $32{\times}32$ patch shuffle); \emph{(b)} the generation stage $t_g\!\in\!\{0,.25,.5,.75\}$; \emph{(c)} the input representation (RGB, the SDXL latent, frozen DINOv2). RGB separability changes sharply with $t$ and is high only near clean, whereas the SDXL latent stays flat and DINOv2 stays separable into the high-noise band. }
    \label{fig:texture}
\end{figure}

Given a Gaussian noise map $\epsilon$ and a conditioning input $c$, a single few-step generator $G_\theta$ produces a clean RGB image $\hat{x}$ in $K$ pixel-space denoising steps (as few as one), with no latent tokenizer or decoder at any point (Figure~\ref{fig:method}). This native RGB output exposes both the matching behavior and the external-guidance opportunity analyzed in Section~\ref{sec:intro}. On the \textbf{teacher-matching side}, converging diagnostics motivate a DMD branch specialized to native RGB through a fixed high-noise \textbf{distribution-matching band} (\textbf{DM-Band}, Section~\ref{sec:dmd}). On the \textbf{real-data side}, direct access to $\hat{x}$ lets one frozen DINOv2 backbone support dual-granularity, critic-decoupled guidance without traversing a VAE decoder: \textbf{DINO-Adv} uses multi-layer spatial patch features (Section~\ref{sec:adv}), whereas \textbf{AF-Loss} uses the normalized final-layer CLS feature for a parameter-free semantic distribution-field objective (Section~\ref{sec:drift}). Only $G_\theta$, the fake-score critic, and the adversarial heads are trained; AF-Loss adds no learnable parameters, and every auxiliary module is discarded at inference.

\subsection{Diagnosing native-RGB teacher matching}\label{sec:dmd}
We distill $G_\theta$ with the DMD objective, using the short schedules $\mathcal{S}_4\!=\!\{0,.25,.5,.75,1\}$ and $\mathcal{S}_1\!=\!\{0,1\}$ ($t\!=\!0$ noise, $t\!=\!1$ clean, DeCo convention). At a matching time $t_m$, we noise the student output once to $x_{t_m}\!=\!\alpha_{t_m}\hat{x}+\sigma_{t_m}\epsilon$ ($\epsilon\!\sim\!\mathcal{N}(0,I)$) and pass this \emph{same} $x_{t_m}$ to both the real score (frozen teacher) and the fake score (a critic initialized from the teacher and trained on noised student samples), which denoise it to clean estimates $x_{0,r}$ and $x_{0,f}$, giving the normalized surrogate gradient
\begin{equation}
g(\hat{x}) = \frac{(\hat{x}-x_{0,r})-(\hat{x}-x_{0,f})}{\operatorname{mean}|\hat{x}-x_{0,r}|+\epsilon},\quad
\mathcal{L}_{\mathrm{DMD}}=\tfrac{1}{2}\|\hat{x}-\operatorname{sg}(\hat{x}-g(\hat{x}))\|_2^2,
\label{eq:dmd}
\end{equation}
with $\operatorname{sg}$ the stop-gradient. The only choice we revisit is which matching levels $t_m\in[.02,.98]$ (lower is noisier) supervise this gradient.

\begin{wrapfigure}{r}{0.46\linewidth}
\centering
\vspace{-2\baselineskip}
\includegraphics[width=\linewidth]{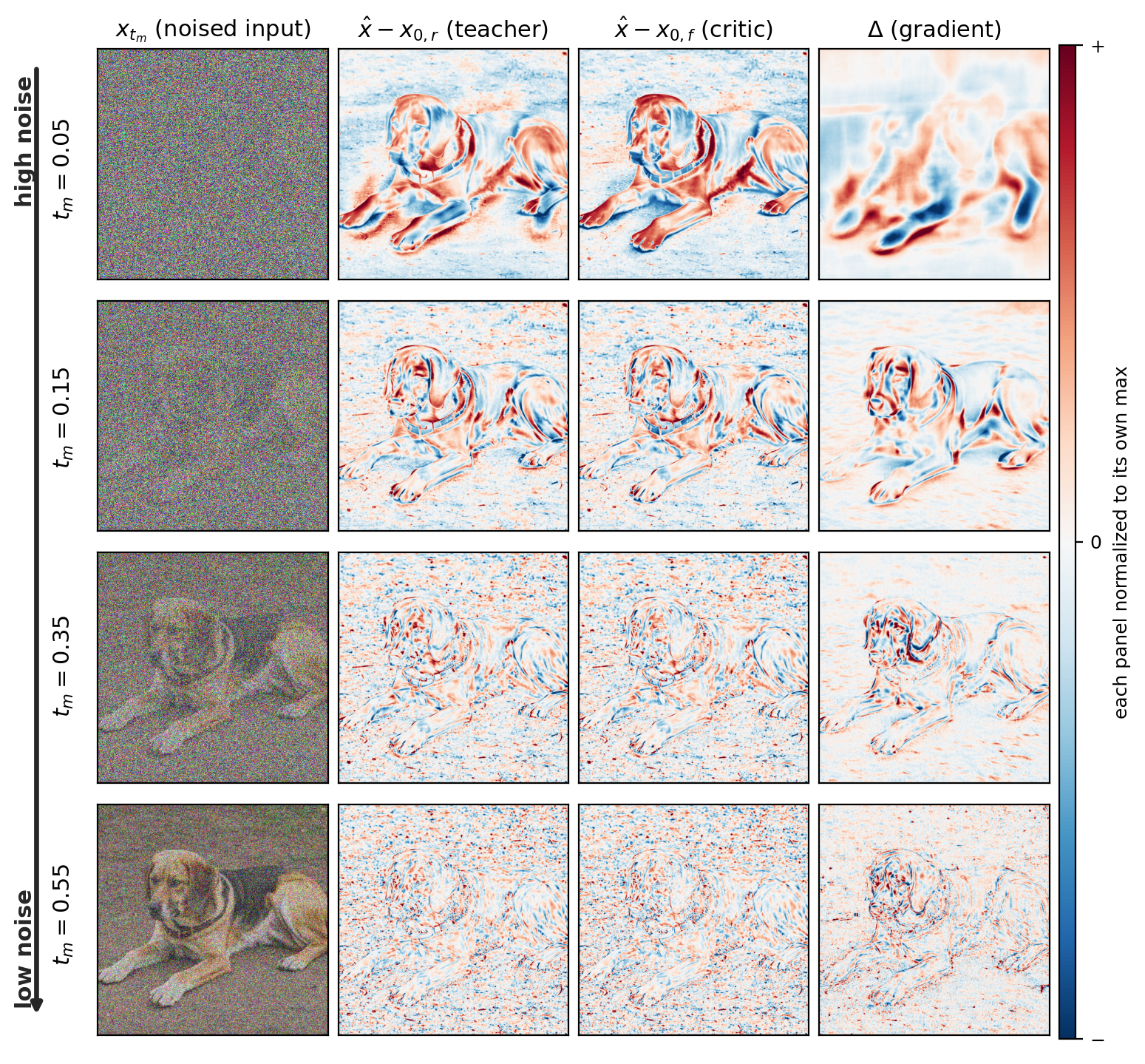}
\vspace{-1\baselineskip}
\caption{Eq.~\ref{eq:dmd} terms across $t_m$; $\Delta$ exhibits broader coherent patterns at high noise and sparse fine-scale patterns near clean. Each panel is independently normalized for spatial visualization; colors are not comparable in magnitude across cells.}
\label{fig:dmdgrad}
\vspace{-1\baselineskip}
\end{wrapfigure}

\paragraph{Analysis: what separates real from generated RGB images.} For our native-RGB teacher, the choice of matching level matters. We therefore probe which cues distinguish real from generated RGB images at each level (Figure~\ref{fig:texture}). Using $1{,}536$ matched real/generated pairs, we noise both members of each pair to the same matching level $t$ and perform a classifier two-sample test (C2ST~\citep{lopezpaz2017c2st}). We fit a linear real/fake classifier on a diagnostic training split and report ROC-AUC on held-out pairs as the separability score ($0.5$ is chance; higher $t$ is cleaner). \emph{(a)} Separability varies sharply with the noise level: high near clean and near chance at high noise, so the two sets are easiest to classify at low noise. The cue survives a high-pass and a $32{\times}32$ patch shuffle but weakens under a low-pass; together, these controls are most consistent with a high-frequency, spatially local texture cue rather than broad structure. \emph{(b)} Across the four generation stages $t_g\!\in\!\{0,.25,.5,.75\}$ the curves nearly coincide, suggesting that the pattern is approximately stage-stable and that one fixed band can serve all stages. \emph{(c)} We compare the classifier's input representation. The latent representation of SDXL remains nearly flat across noise levels, whereas RGB and DINOv2 features vary substantially with noise. RGB texture separability is concentrated near clean images; in contrast, semantic information represented by DINOv2~\citep{oquab2023dinov2} remains discriminative between real and generated samples into the high-noise range. We use C2ST to localize the noise regions in which real and generated representations differ. This C2ST localizes \emph{where} a real-versus-generated cue exists; the update-level and causal evidence come from the teacher--critic difference $\Delta$ (Figure~\ref{fig:dmdgrad}) and the band intervention with its latent negative control. Frequency and spatial controls further indicate that the near-clean separable cue is predominantly local and high-frequency; the gradient visualization below illustrates its spatial form, while the matching-band intervention and latent-space negative control further test the resulting native-RGB matching hypothesis.

\vspace{-0.8\baselineskip}
\paragraph{The high-noise band (DM-Band).}
To preserve DMD's core goal of constraining structure and semantics, we keep the near-clean texture cue out of the gradient by truncating the matching range and supervising DMD only at high noise. In Figure~\ref{fig:texture}c, for $t\!\lesssim\!.35$, the semantics-oriented DINOv2 features distinguish real from generated samples, whereas the texture- and detail-sensitive RGB representation remains weakly separable. Figure~\ref{fig:dmdgrad} visualizes the Eq.~\ref{eq:dmd} terms (the noised input, the teacher and critic residuals, and their difference $\Delta$), each panel scaled to its own maximum: this visualization compares spatial content, not magnitude. The gradient $\Delta$ exhibits broad coherent patterns at $t_m\!=\!.35$ and sparse fine-scale patterns by $t_m\!=\!.55$, placing $.35$ near the observed transition. Combining the two diagnostics, we draw matching levels from $t_m\!\sim\!U(.02,.35)$, applied to the DMD branch only. The band ablation below then tests this targeted intervention, while the latent-space negative control tests whether the intervention is specific to the observed native-RGB matching bias. It is a single hand-set hyperparameter with no stage-dependent schedule, classifier at training time, or online estimation.

\vspace{-1.0\baselineskip}
\paragraph{Latent-space negative control.}
\begin{wraptable}{r}{0.42\linewidth}
\vspace{-0.6\baselineskip}
\centering\small
\setlength{\tabcolsep}{5pt}
\begin{tabular}{@{}lcc@{}}
\toprule
\dpg$\uparrow$ & $U(.02,.98)$ & $U(.02,.35)$ \\
\midrule
Pixel (DeCo) & 82.55 & \textbf{82.96} \\
Latent (SDXL) & \textbf{67.64} & 63.23 \\
\bottomrule
\end{tabular}
\vspace{-1.1\baselineskip}
\end{wraptable}
Because the prescription is motivated by the observed near-clean RGB texture bias, it should not be universally beneficial when that representation-level bias is absent. On latent SDXL-DMD2 the same band lowers DPG ($67.64\!\to\!63.23$), opposite to native RGB. This is consistent with VAE compression suppressing low-level cues, making DM-Band a native-RGB remedy rather than a universal DMD schedule.

\subsection{Critic-decoupled local adversary (DINO-Adv)}\label{sec:adv}
Matching the teacher alone is limited by its imperfect score on real data~\citep{yin2024dmd2}, so we add guidance that learns from real images directly. We exploit a design freedom exposed by native RGB outputs: because the student emits clean RGB natively, real-data guidance can be evaluated directly by an external pretrained model without traversing a VAE decoder or sharing the fake-score critic. DINO-Adv decouples the adversarial gradient path from the co-adapting fake-score critic with $560$M learnable parameters and instead uses an $86$M-parameter frozen DINOv2 ViT-B/14 backbone with only $1.58$M learnable head parameters, reducing per-step training time from $1.82$ to $0.72$ seconds ($2.5\times$ faster) while supplying a stable local realism signal.
Specifically, we independently sample a noise level for each real image \(x\) and generated image \(\hat{x}\), and feed the resulting noisy RGB images into a frozen DINOv2 ViT-B/14; layers $\{2,5,8,11\}$ provide spatial patch features $f_l$ to learnable heads
\begin{equation}
h_l(f_l)=\mathrm{Conv}_{1\times1}^{768\rightarrow512}\!\rightarrow\mathrm{GN}_{32}\!\rightarrow\mathrm{LeakyReLU}\!\rightarrow\mathrm{Conv}_{1\times1}^{512\rightarrow1},
\end{equation}
which output patch logits, trained with the non-saturating softplus objective. The same frozen backbone later supplies the normalized final-layer CLS feature for AF-Loss (Section~\ref{sec:drift}), so one representation supports both local parametric patch guidance and a parameter-free semantic distribution-field objective. The default weight is $\lambda_{\mathrm{GAN}}=.01$.

\paragraph{Effect of critic decoupling.}
\begin{wraptable}{r}{0.47\linewidth}
\raggedleft\scriptsize
\vspace{-0.9\baselineskip}
\setlength{\tabcolsep}{3pt}
\resizebox{\linewidth}{!}{
\begin{tabular}{@{}lccccrr@{}}
\toprule
 & \multicolumn{2}{c}{\dpg\,$\uparrow$} & \multicolumn{2}{c}{\geneval\,$\uparrow$} & \multicolumn{2}{c}{Train cost\,$\downarrow$} \\
\cmidrule(lr){2-3}\cmidrule(lr){4-5}\cmidrule(lr){6-7}
Adversary & $1$ & $4$ & $1$ & $4$ & s/step & \shortstack{\#learn.\\[-1pt]\tiny(adv. path)}\\
\midrule
DMD2 Feat.\ GAN & 78.35 & 82.12 & 0.6960 & 0.7709 & 1.82 & 560M \\
Frozen-DINOv2 & \textbf{81.26} & \textbf{82.55} & \textbf{0.7723} & \textbf{0.8155} & \textbf{0.72} & \textbf{1.58M} \\
\bottomrule
\end{tabular}
}
\end{wraptable}
We compare the two discriminators head-to-head by porting DMD2's feature GAN unchanged (a head on block-$7$ of the $16$-block critic, the ``Feature GAN'' baseline) and swapping only that critic-coupled head for the frozen-DINOv2 adversary, keeping the same DMD and schedule. The frozen-DINOv2 adversary comprises an $86$M-parameter frozen ViT-B/14 backbone and only $1.58$M learnable head parameters, whereas the DMD2 Feature GAN uses $560$M learnable adversary parameters. The swap improves both \dpg\ and \geneval\ at one and four steps, reaching $0.8155$ four-step \geneval, with the gap widening as the trajectory shortens, and it trains about $2.5\times$ faster ($0.72$ vs.\ $1.82$\,s/step).

\subsection{Parameter-free semantic distribution anchor-field loss (AF-Loss)}
\label{sec:drift}

\begin{figure}[t]
    \centering
    \includegraphics[width=\linewidth]{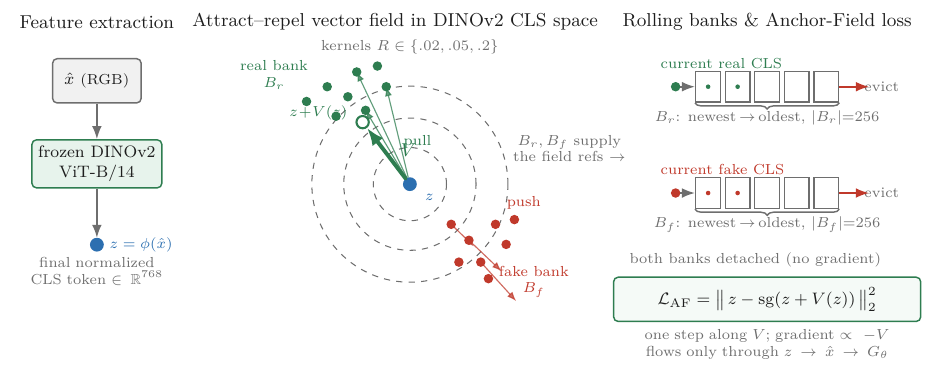}
    \caption{\textbf{Anchor-Field Loss (AF-Loss).} The generated image's normalized final-layer DINOv2 CLS feature $z=\phi(\hat{x})$ is attracted toward a rolling bank of real features $B_r$ and repelled from a bank of generated features $B_f$, giving a displacement $V(z)$. AF-Loss converts this displacement into a stopped target whose gradient flows only through $z\!\to\!\hat{x}\!\to\!G_\theta$; the field itself has no learnable parameters. The banks are detached FIFO queues capped at $256$ features.}
    \label{fig:drift}
\end{figure}
DINO-Adv supplies local parametric patch guidance, but its patch decisions do not specify a distribution-level direction in semantic feature space. A complementary objective for DMD should provide such a direction without replacing the conditional teacher objective, learning another estimator, or modifying inference. We therefore introduce AF-Loss, a \emph{parameter-free semantic distribution anchor-field loss for DMD}. It turns detached empirical real and generated feature supports into a sample-dependent, multi-scale displacement and injects that displacement as a stopped target for the generator. This is distinct from direct feature regression, whose target does not depend on the geometry of both empirical supports, and from standalone field-based generation~\citep{deng2026drifting}, where the field itself defines the generative process.

Concretely, AF-Loss operates on normalized final-layer DINOv2 CLS features extracted directly from generated RGB. Multi-scale kernels attract each generated feature toward real-feature neighborhoods and repel it from generated-feature neighborhoods. DMD continues to provide teacher and prompt guidance, while AF-Loss reuses the same frozen encoder as DINO-Adv and adds no learnable parameters or inference-time computation. For a stable empirical support, current real and student features are prepended to two detached FIFO banks $B_r$ and $B_f$, each capped at $256$ features; the sole trainable path is $z\!\to\!\hat{x}\!\to\!G_\theta$ (Figure~\ref{fig:drift}).

We measure similarity in the DINOv2 feature space with the Euclidean distance $d(z,u)=\|z-u\|$, normalized by the batch-mean distance $\bar{d}$ for scale invariance. At bandwidth $R$ the affinity is a softmax kernel $K_R(z,u)\propto\exp\!\big(\!-d(z,u)/(\bar{d}\,R)\big)$, symmetrized over the batch and self-masked so a sample never interacts with its own copy in $B_f$. Attracting $z$ toward its real neighbors in $B_r$ and repelling it from its generated neighbors in $B_f$, kernel-weighted and summed over the bandwidths $R\in\mathcal{R}$, gives an illustrative simplified field $\widetilde{V}(z)$ at $z=\phi(\hat{x})$
\begin{equation}
\widetilde{V}(z)=\sum_{R\in\mathcal{R}}\left[
\frac{\sum_{r\in B_r}K_R(z,r)(r-z)}{\sum_{r\in B_r}K_R(z,r)+\epsilon}
-\frac{\sum_{f\in B_f}K_R(z,f)(f-z)}{\sum_{f\in B_f}K_R(z,f)+\epsilon}\right].
\end{equation}
$\widetilde{V}$ is an illustrative single-softmax simplification; the actual field $V$ used below is the symmetrized (doubly-normalized), cross-coupled, per-scale force-normalized field summed over $\mathcal{R}$ and defined exactly in Appendix~\ref{app:afimpl} (Algorithm~\ref{alg:af}). Because $V$ is treated as a fixed target displacement (stop-gradient), the loss simply moves $z$ one step along the field,
\begin{equation}
\mathcal{L}_{\mathrm{AF}}=\|\phi(\hat{x})-\operatorname{sg}(\phi(\hat{x})+V(\phi(\hat{x})))\|_2^2,
\end{equation}
so its gradient with respect to $z$ points along $-V(z)$, up to the positive scalar induced by the squared-error reduction, and the signal flows only through $z\!\to\!\hat{x}\!\to\!G_\theta$. The reference clouds are \emph{detached} first-in-first-out queues: at each step we prepend the current real and student features and truncate to $N_b\!=\!256$,
\begin{equation}
B_r\leftarrow[\,\phi(x)\,;\,B_r\,]_{1:N_b},\qquad B_f\leftarrow[\,\phi(\hat{x})\,;\,B_f\,]_{1:N_b},
\label{eq:bank}
\end{equation}
With a global batch of $32$, each step admits the newest $32$ real and $32$ generated tokens and evicts the oldest $32$, so every $256$-token cloud spans the current batch and the previous seven; the clouds receive no gradient. The balanced anchor uses $|B_r|=|B_f|=N_b=256$, $\mathcal{R}=\{.02,.05,.2\}$, and $\lambda_{\mathrm{AF}}=.05$; our maximum-DPG variant changes only the Anchor-Field weight to $.10$ and the radii to $\{.01,.05,.5\}$. No other auxiliary losses are used. Kernel batch symmetrization, the scope of $\bar d$, and the loss reduction are detailed in Appendix~\ref{app:afimpl}. The total generator objective is
\begin{equation}
\mathcal{L}_G=\mathcal{L}_{\mathrm{DMD}}+\lambda_{\mathrm{GAN}}\mathcal{L}_{\mathrm{GAN}}+\lambda_{\mathrm{AF}}\mathcal{L}_{\mathrm{AF}}.
\end{equation}

\begin{figure*}[t]
\centering
\setlength{\tabcolsep}{1pt}
\begin{tabular}{@{}p{0.487\linewidth}@{\hspace{4pt}}p{0.487\linewidth}@{}}
\qmh & \qmh \\[2pt]
\qpl{A set of four green plastic food containers displayed against a stark white background, each captured from a distinct angle to showcase the varying perspectives\,\ldots}
& \qpl{An up-close image showcasing the intricate interior of a walnut, split cleanly down the middle to reveal its textured, brain-like halves\,\ldots} \\
\qi{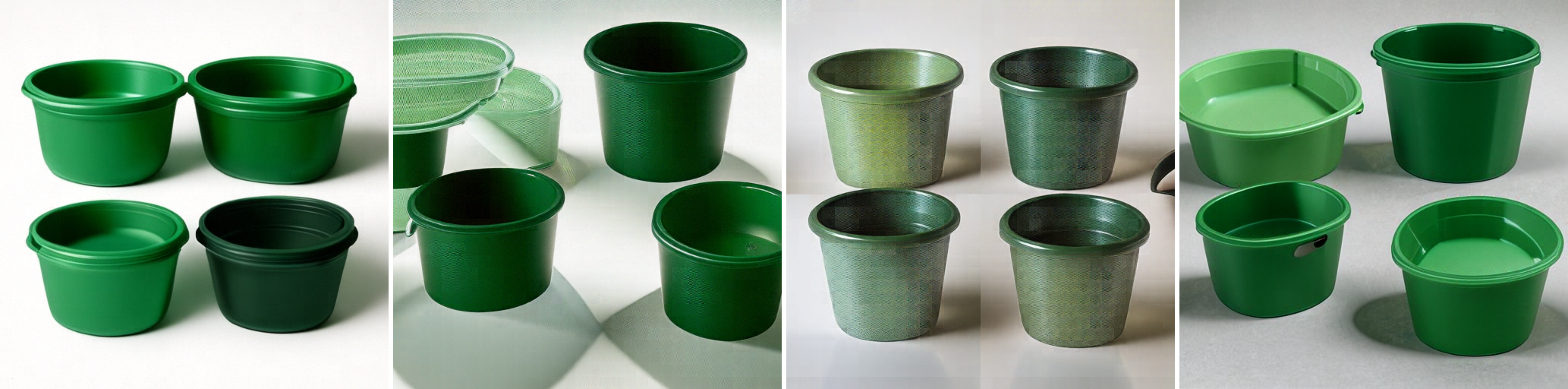} & \qi{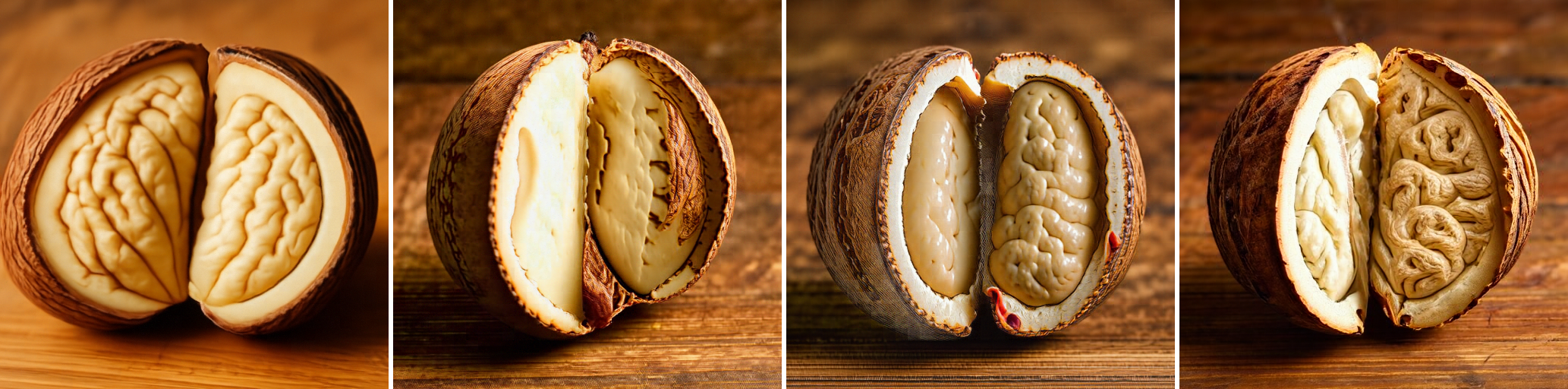} \\[5pt]
\qpl{Three sleek, dark wooden boats are resting along the banks of a tranquil, azure blue lake, their oars tucked neatly inside\,\ldots}
& \qpl{A sprawling field blanketed with vibrant wildflowers, where a tall giraffe and a striped zebra stand side by side under acacia trees\,\ldots} \\
\qi{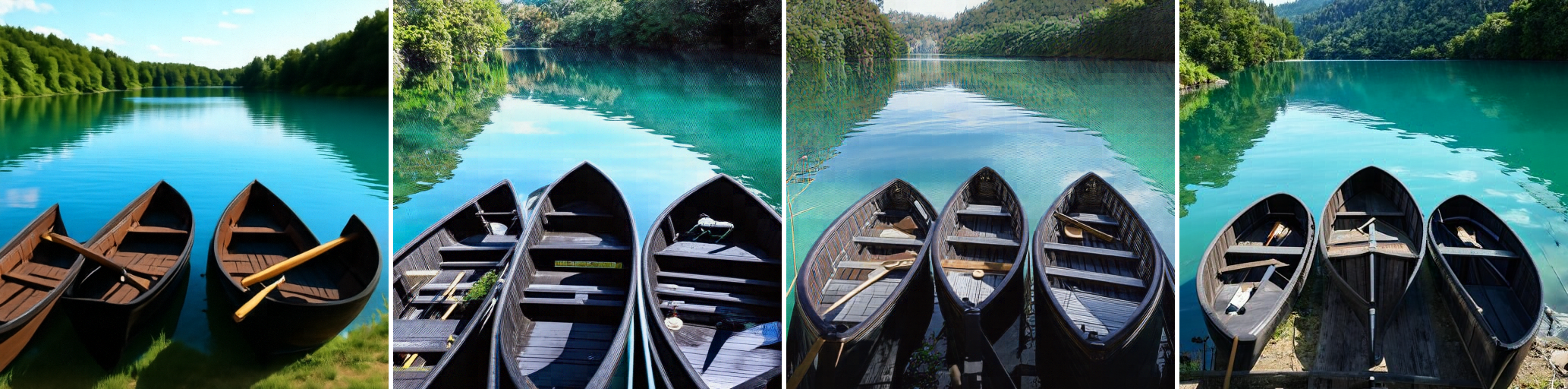} & \qi{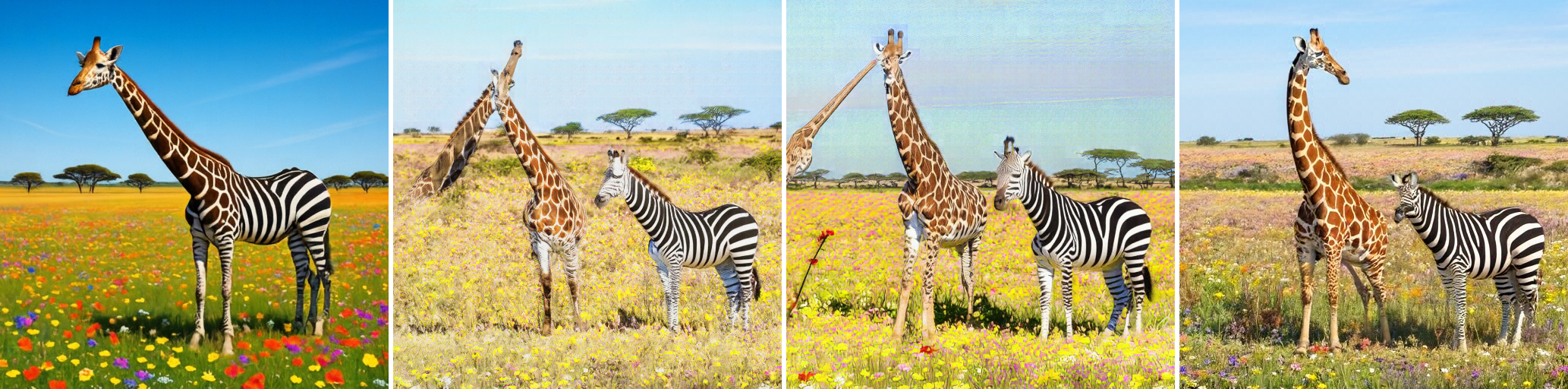} \\[7pt]
\qmh & \qmh \\[2pt]
\qps{A cat on a leather chair next to remotes.} & \qps{A large dog sitting on top of a roof.} \\
\qi{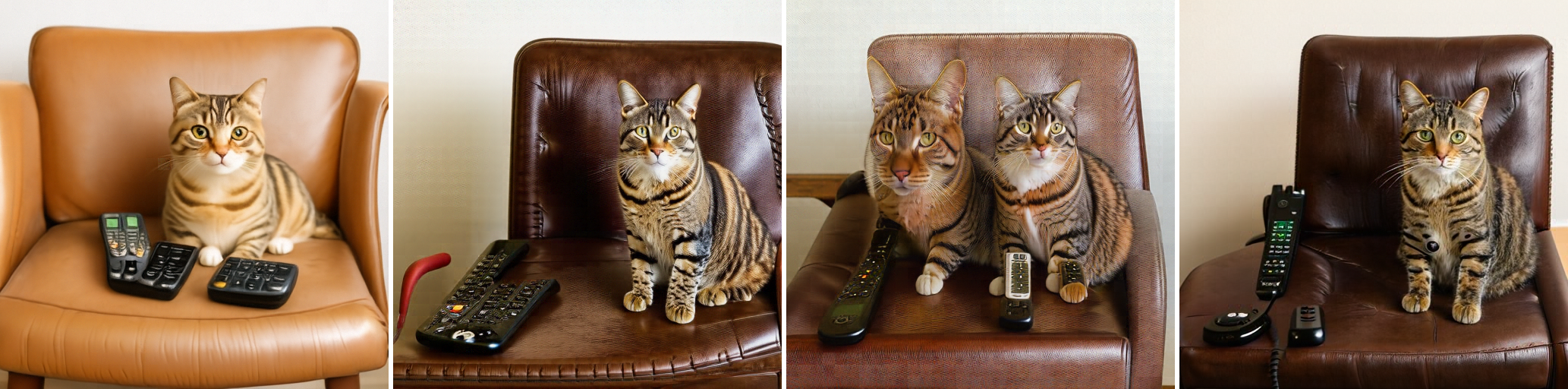} & \qi{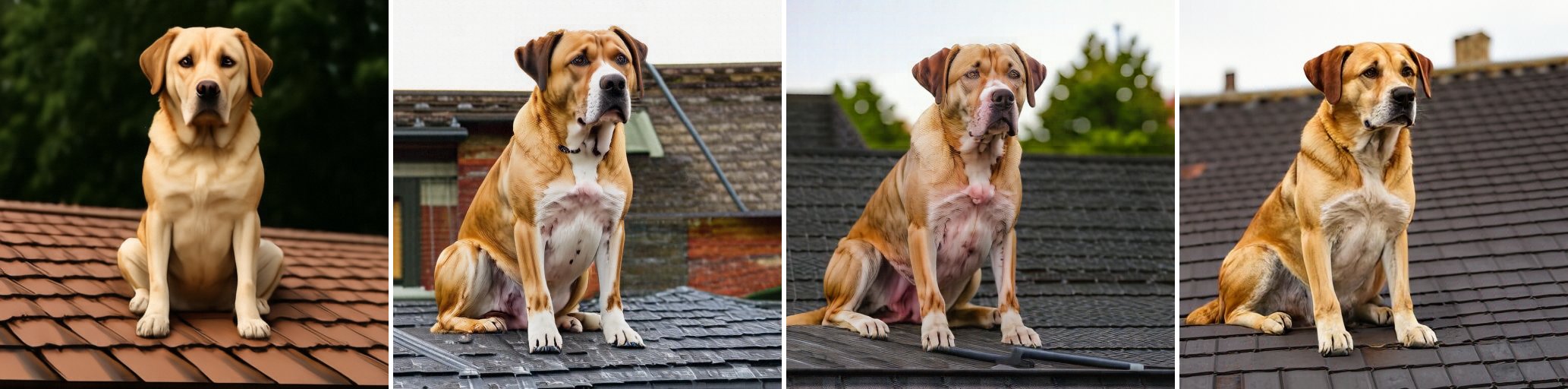} \\[5pt]
\qps{A gray cat standing on top of a black car.} & \qps{A zebra standing in a field under the shade of a nearby tree.} \\
\qi{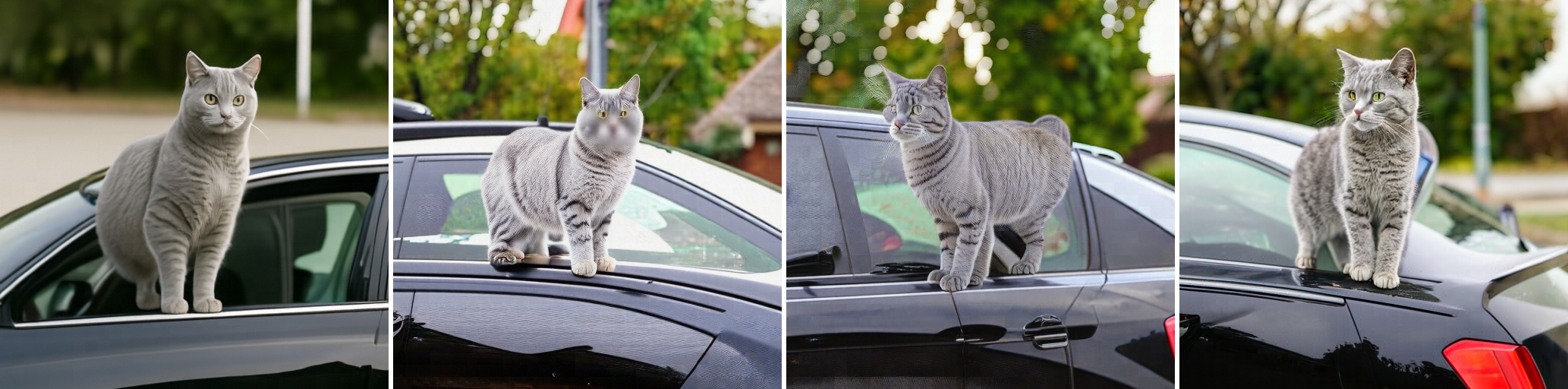} & \qi{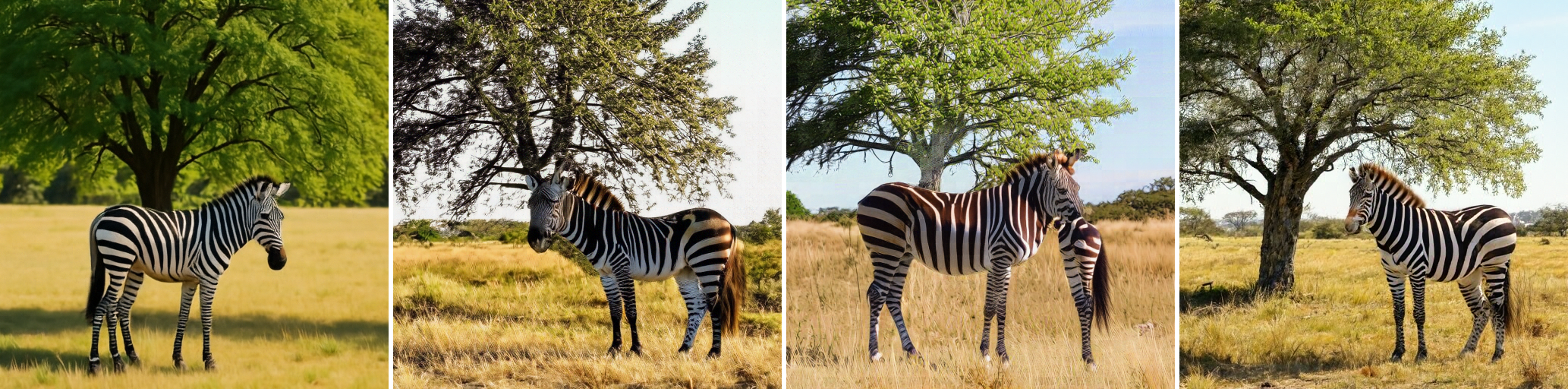} \\
\end{tabular}
\caption{Qualitative comparison at four steps. Top block: long, densely specified \dpg\ prompts (abbreviated with ``\ldots''); bottom block: short COCO captions. \method{} preserves object count and composition and recovers fine structure, across both long and short text.}
\label{fig:qual}
\end{figure*}

\section{Experiments}
\subsection{Experimental setting}
\paragraph{Implementation details.} We distill a 1.1B-parameter, 25-step DeCo teacher \citep{ma2025deco} at $512\times512$ resolution. The student is initialized from the teacher and trained for 20k optimizer steps on eight GPUs with batch size 32. Optimization uses AdamW without weight decay, at learning rates $2\times10^{-6}$ (generator), $5\times10^{-6}$ (fake-score network), and $2\times10^{-4}$ (adversarial heads), with the fake-score network updated five times per generator step; classifier-free guidance is 4.0. 
The canonical \method{} student combines the fixed high-noise band $U(.02,.35)$ with dual-granularity real-data guidance over one frozen DINOv2 encoder: local parametric DINO-Adv at weight $0.01$ and parameter-free semantic distribution-field AF-Loss at weight $0.05$, using radii $\{.02,.05,.2\}$ and 256-token real/generated banks. The four-step generator denoises at $t\!\in\!\{0,.25,.5,.75\}$. 

\paragraph{Dataset and benchmark.} Training uses $200$k image--caption pairs from the BLIP3-o long-caption split \citep{chen2025blip3o}, which pairs open-domain images with detailed Qwen-generated captions. We evaluate prompt following with \dpg~\citep{hu2024ella} (1{,}065 readable 2-by-2 grids, four images per prompt, 4{,}260 images total) and compositionality on \geneval~\citep{ghosh2023geneval} (553 prompts, four samples each, 2{,}212 images). Prompt--image alignment additionally uses VQAScore~\citep{lin2024vqascorer} on the full 1{,}600-prompt GenAI-Bench~\citep{li2024genaibench}, with one generated image per prompt. Fidelity and no-reference quality use COCO30K with 30{,}000 readable $512^2$ COCO samples, reporting IS, CLIP similarity, recall, and NIQE.

\paragraph{Compared methods.} We compare against some few-step methods, each re-implemented on the same pixel-space DeCo teacher and trained under an identical 20k-step budget and shared protocol: Flash~\citep{chadebec2025flash}, DMD2~\citep{yin2024dmd2}, TDM~\citep{luo2025tdm}, Decoupled DMD~\citep{decoupleddmd2026}, and DMDR~\citep{jiang2026dmdr}. All compared methods are trained with three independent seeds, and report the mean. Objective-specific hyperparameters follow the corresponding papers, while the data, optimizer budget, and evaluation protocol are held fixed.

\begin{table}[t]
\centering
\caption{Main comparison. All few-step methods are re-implemented on the same teacher; bold/underline mark the best/second-best students at each step count.}
\label{tab:main}
\scriptsize
\setlength{\tabcolsep}{2.4pt}
\resizebox{\textwidth}{!}{%
\begin{tabular}{@{}lcrrrrrrr@{}}
\toprule
Model & NFE & \geneval$\uparrow$ & DPG$\uparrow$ & VQA$\uparrow$ & IS$\uparrow$ & CLIP$\uparrow$ & Rec.$\uparrow$ & NIQE$\downarrow$ \\
\midrule
DeCo~\citep{ma2025deco} (official) & 25 & 0.8216 & 81.580 & 0.7014 & 35.58 & 0.3200 & 0.3110 & 4.064 \\
\midrule
Flash~\citep{chadebec2025flash} & 4 & 0.7814 & 76.703 & 0.6900 & 29.94 & 0.3178 & 0.1486 & 5.688 \\
DMD2~\citep{yin2024dmd2} & 4 & 0.7709 & 82.115 & 0.7058 & \underline{36.76} & 0.3170 & 0.4349 & 4.168 \\
TDM~\citep{luo2025tdm} & 4 & 0.7924 & 82.367 & 0.7071 & 36.48 & 0.3165 & \underline{0.4447} & 3.871 \\
Decoupled~\citep{decoupleddmd2026} & 4 & \underline{0.8100} & 83.095 & \underline{0.7111} & 36.51 & 0.3195 & 0.3516 & 3.953 \\
DMDR~\citep{jiang2026dmdr} & 4 & 0.8063 & \underline{83.160} & 0.7023 & 34.22 & \underline{0.3198} & 0.4414 & \underline{3.710} \\
\method{} (ours) & 4 & \textbf{0.8232} & \textbf{83.692} & \textbf{0.7146} & \textbf{41.01} & \textbf{0.3203} & \textbf{0.4701} & \textbf{2.925} \\
\midrule
Flash~\citep{chadebec2025flash} & 1 & 0.6104 & 68.438 & 0.6517 & 28.74 & 0.3147 & 0.0490 & 5.103 \\
DMD2~\citep{yin2024dmd2} & 1 & 0.6960 & 78.352 & 0.6972 & 34.43 & 0.3196 & \underline{0.2834} & \underline{3.489} \\
TDM~\citep{luo2025tdm} & 1 & 0.7129 & 79.267 & 0.7043 & 34.87 & 0.3222 & 0.2785 & 3.672 \\
Decoupled~\citep{decoupleddmd2026} & 1 & \underline{0.7826} & 80.792 & 0.7044 & \underline{36.55} & \underline{0.3236} & 0.2639 & 3.562 \\
DMDR~\citep{jiang2026dmdr} & 1 & 0.7512 & \underline{82.395} & \underline{0.7076} & 34.15 & 0.3229 & 0.2598 & 3.973 \\
\method{} (ours) & 1 & \textbf{0.7914} & \textbf{82.963} & \textbf{0.7155} & \textbf{38.99} & \textbf{0.3243} & \textbf{0.3447} & \textbf{3.409} \\
\bottomrule
\end{tabular}}
\end{table}

\subsection{Comparison with existing methods}
Table~\ref{tab:main} compares \method{} with five distillation baselines and the official $25$-step teacher. The four-step model ranks first among the evaluated distillers on all seven metrics: it reaches $83.692$ \dpg\ and $0.8232$ \geneval, ahead of the strongest corresponding baselines (DMDR at $83.160$ \dpg\ and Decoupled at $0.8100$ \geneval), and raises VQAScore to $0.7146$. On COCO30K it also obtains the strongest IS ($41.01$), recall ($0.4701$), and NIQE ($2.925$). The same advantage persists at one step, indicating that the gains are not tied to a single trajectory length.

The efficiency improvement is measured under a common training pipeline, hardware, and global batch size: \method{} takes $0.72$\,s/step, compared with $0.90$ for DMDR and $1.82$ for DMD2 and Decoupled DMD, making it about $2.5\times$ faster than the latter two (Appendix Table~\ref{tab:training_speed}). Figure~\ref{fig:qual} further shows improved object count, composition, and fine structure on both long and short prompts. The recipe also transfers to PixelGen-XXL, which performs better than its teacher (Appendix~\ref{app:pixelgen}).

\subsection{Ablation studies}
\label{sec:ablation}
\begin{figure}[t]
    \centering
    \includegraphics[width=\linewidth]{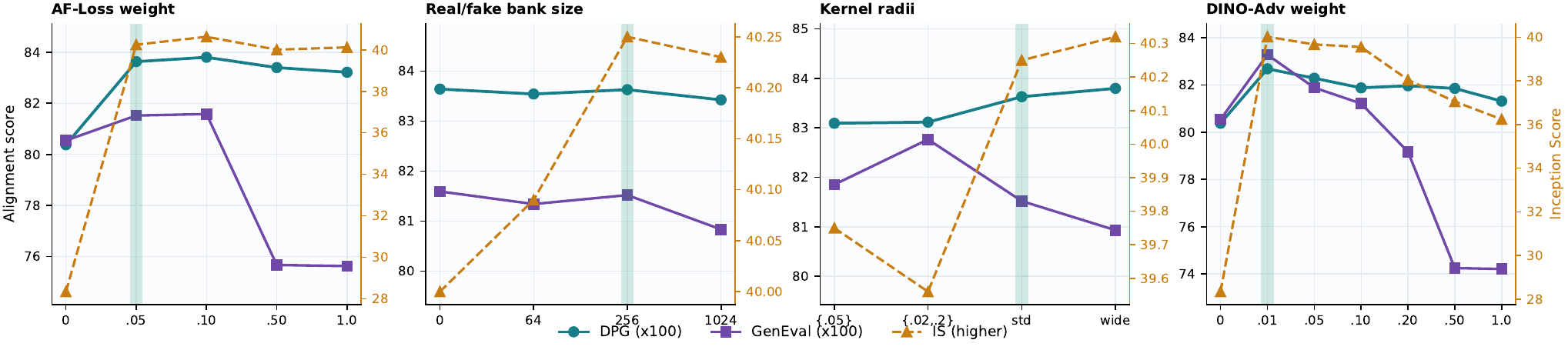}
    \caption{Hyperparameter sensitivity in DPG, \geneval, and IS; green bands mark defaults. These are OFAT sweeps around an anchor that is not the full model, so their default and zero-weight points are not directly comparable to the Table~\ref{tab:ablation} component ablation (anchor protocol in Appendix~\ref{app:affeat}).}
    \label{fig:ofat}
\end{figure}

We study the contribution of each signal in Table~\ref{tab:ablation} and sensitivity to the four main hyperparameters in Figure~\ref{fig:ofat}. Appendix~\ref{app:band} analyzes matching-band selection and the diagnostic that motivates it. All ablations use four-step students under the main-comparison protocol.

\paragraph{Component contribution.} Table~\ref{tab:ablation} separates DM-Band from the two RGB-native real-data objectives. DINO-Adv raises \dpg\ from $81.1$ to $82.6$ and IS from $34.2$ to $39.4$, while AF-Loss alone gives the strongest single-objective \dpg\ ($83.2$). Their combination without DM-Band reaches $83.427$ \dpg\ and $0.7113$ VQAScore, showing that AF-Loss complements the local adversary. Replacing AF-Loss with direct DINOv2-CLS regression degrades six of seven metrics, including $1.12$ DPG and $1.61$ IS (Appendix~\ref{app:affeat}); thus the gain is not explained by frozen features alone. DM-Band is complementary: on the complete real-data branch it raises \geneval\ from $0.8178$ to $0.8232$ and IS from $39.52$ to $41.01$, while lowering NIQE from $3.307$ to $2.925$. 

\begin{table}[t]
\centering
\caption{Component ablation. Bold/underline mark the best/second-best results.}
\label{tab:ablation}
\small
\setlength{\tabcolsep}{5pt}
\begin{tabular*}{\textwidth}{@{\extracolsep{\fill}}cccrrrrrr@{}}
\toprule
\multicolumn{3}{c}{Module} & \multicolumn{6}{c}{Metric} \\
\cmidrule(lr){1-3}\cmidrule(lr){4-9}
DM-Band & DINO-Adv & AF-Loss & \geneval$\uparrow$ & DPG$\uparrow$ & VQA$\uparrow$ & IS$\uparrow$ & Rec.$\uparrow$ & NIQE$\downarrow$ \\
\midrule
 & & & 0.8037 & 81.115 & 0.6930 & 34.24 & 0.2802 & 5.174 \\
 & \checkmark & & 0.8155 & 82.550 & 0.7045 & 39.38 & 0.4297 & 3.337 \\
\checkmark & \checkmark & & \underline{0.8193} & 82.959 & 0.7077 & \underline{40.21} & 0.4318 & \underline{3.034} \\
 & & \checkmark & 0.8090 & 83.194 & 0.7079 & 39.46 & 0.3161 & 3.574 \\
 & \checkmark & \checkmark & 0.8178 & \underline{83.427} & \underline{0.7113} & 39.52 & \underline{0.4324} & 3.307 \\
\checkmark & \checkmark & \checkmark & \textbf{0.8232} & \textbf{83.692} & \textbf{0.7146} & \textbf{41.01} & \textbf{0.4701} & \textbf{2.925} \\
\bottomrule
\end{tabular*}
\end{table}

\paragraph{Hyperparameter ablation.} Figure~\ref{fig:ofat} reports one-at-a-time sweeps in DPG, \geneval, and IS. Increasing $\lambda_{\mathrm{AF}}$ from zero to $.05$ yields the largest improvement, whereas stronger weights sharply reduce \geneval. A bank size of $256$ gives the highest IS while leaving DPG and \geneval\ stable; kernel-radius variants are comparatively insensitive, so we retain $\{.02,.05,.2\}$. Finally, $\lambda_{\mathrm{GAN}}\!=\!.01$ gives the best joint result among the tested adversarial weights.

\vspace{-0.5em}
\section{Conclusion}
\vspace{-0.5em}

We presented \method{}, which specializes DMD's teacher matching and real-data guidance for native-RGB generation. Converging diagnostics identify a near-clean local-texture bias that motivates DM-Band. Native RGB then enables direct external supervision: DINO-Adv decouples local patch guidance from the fake-score critic, while AF-Loss adds parameter-free auxiliary semantic distribution-field guidance, operating on detached rolling supports alongside prompt-conditioned teacher supervision with no learned estimator or inference-time cost. Across four benchmarks, the four-step student exceeds its teacher on all seven metrics and ranks first among the evaluated distillers.

\bibliography{references}
\bibliographystyle{assets/plainnat}

\newpage
\appendix

\section{Implementation and Reproducibility Details}
\label{app:implementation}
\paragraph{Optimization.}
Training uses eight data-parallel ranks, per-rank batch size 4, and 20k optimizer steps. The generator, fake-score network, and DINOv2 heads use separate optimizers. The generator-side update ratio is 5. Checkpoints are recorded at 5k increments. The DINOv2 backbone remains in evaluation mode and is never updated.

\paragraph{Timing protocol and parameter accounting.}
One \emph{step} denotes a single generator optimizer update, so ``20k steps'' counts generator updates; the fake-score critic is updated five times per generator step (update ratio 5). Reported seconds-per-step are end-to-end wall-clock over one generator step \emph{including} its five critic updates and all loss terms, measured on eight A100-80GB GPUs in bf16 autocast mixed precision after 50 warm-up steps, averaged over 200 steps with device synchronization. The $1.58$M figure is the learnable size of the adversarial heads only: the $560$M fake-score critic is trained by every method, so the speedup does not come from fewer trainable parameters but from removing the adversarial gradient path through that heavy critic. DMD2's feature-GAN back-propagates its discriminator through the $560$M critic, whereas DINO-Adv attaches a $1.58$M head to a frozen $86$M DINOv2. Here ``frozen'' means we do not compute or apply gradients to the encoder weights; the input-gradient path through $\phi(\hat x)$ to the image, and hence to $G_\theta$, is retained, as the generator update requires.

\paragraph{Same-pipeline training speed.}
Table~\ref{tab:training_speed} reports end-to-end optimizer-step time measured under the same training pipeline, hardware, and global batch size. DMD2 and Decoupled each require $1.82$ seconds per step, DMDR requires $0.90$ seconds, and \method{} requires $0.72$ seconds per step.
\begin{table}[h]
\centering
\caption{Same-pipeline training speed. All methods use the same hardware, global batch size, and end-to-end step-timing protocol. Lower is better.}
\label{tab:training_speed}
\small
\setlength{\tabcolsep}{12pt}
\begin{tabular}{@{}lc@{}}
\toprule
Method & Time (s/step)$\downarrow$ \\
\midrule
DMD2~\citep{yin2024dmd2} & 1.82 \\
Decoupled DMD~\citep{decoupleddmd2026} & 1.82 \\
DMDR~\citep{jiang2026dmdr} & 0.90 \\
\textbf{\method{} (ours)} & \textbf{0.72} \\
\bottomrule
\end{tabular}
\end{table}

\paragraph{Overlap diagnostic.}
We use 1,536 matched real/generated pairs and five grouped splits. DINOv2 layers $\{2,5,8,11\}$ are mean-pooled and concatenated. Separability is measured in matching-time increments of .05 across early, middle, and late generator checkpoints and stages $t_g\in\{0,.25,.5,.75\}$, showing that the noise-dependent separability pattern is essentially shared across generation stages (Section~\ref{sec:dmd}).

\paragraph{Exact evaluation.}
DPG acceptance requires 4,260 generated images, 1,065 readable 2-by-2 grids, and zero skipped grids. GenAI-Bench acceptance requires 1,600 readable images, one for each prompt, scored with VQAScore. COCO30K acceptance requires 30,000 readable $512\times512$ images and zero corrupt files. \geneval\ acceptance requires $553\times4=2,212$ images scored with the original DeCo evaluator. The result registry closes an endpoint only when all required metrics and the exact source revision are recorded.

\section{Extension to another backbone: PixelGen}
\label{app:pixelgen}
To test whether the recipe transfers beyond DeCo, we apply the same \method{} distillation to a PixelGen-XXL teacher and evaluate few-step students against it. Table~\ref{tab:pixelgen} reports the same alignment and quality metrics as the main comparison. The PixelGen-XXL results show the same trend: the four-step student performs better than its $25$-step teacher on every reported metric (e.g., DPG $79.360\!\to\!81.708$, \geneval\ $0.7965\!\to\!0.8112$, IS $37.24\!\to\!40.46$), and the one-step student still stays ahead on dense-prompt following, VQAScore, and CLIP, providing evidence that the native-RGB matching band, DINOv2 adversary, and Anchor-Field loss are not tied to a single backbone.
\begin{table}[H]
\centering
\caption{Extension to a PixelGen-XXL teacher: \method{} few-step students vs.\ the original PixelGen teacher. Metrics as in Table~\ref{tab:main}; $\uparrow$/$\downarrow$ give direction, \textbf{bold} is best and \underline{underline} second best per column. The four-step student uses the strictly-rerun schedule-matched replica.}
\label{tab:pixelgen}
\small
\setlength{\tabcolsep}{5pt}
\begin{tabular}{@{}lcccccccc@{}}
\toprule
Model & NFE & \geneval$\uparrow$ & DPG$\uparrow$ & VQA$\uparrow$ & IS$\uparrow$ & CLIP$\uparrow$ & Rec.$\uparrow$ & NIQE$\downarrow$ \\
\midrule
PixelGen-XXL teacher & 25 & 0.7965 & 79.360 & 0.6804 & 37.24 & 0.3141 & 0.3328 & 4.064 \\
\method{}-PixelGen (ours) & 4 & \textbf{0.8112} & \textbf{81.708} & \underline{0.6897} & \textbf{40.46} & \underline{0.3184} & \textbf{0.4892} & \textbf{3.088} \\
\method{}-PixelGen (ours) & 1 & \underline{0.8072} & \underline{80.898} & \textbf{0.6961} & \underline{39.48} & \textbf{0.3232} & \underline{0.3872} & \underline{3.275} \\
\bottomrule
\end{tabular}
\end{table}

\section{Noise-band selection and separability}
\label{app:band}
This appendix supports the fixed high-noise matching band of Section~\ref{sec:dmd} with the separability diagnostic that motivates it and the full band-position sweep.

\paragraph{Matching-band position.} Table~\ref{tab:bandpos} moves the DMD matching range across noise while holding everything else fixed. The tight high-noise band $U(.02,.35)$ gives the best prompt following ($82.959$ \dpg) and compositionality ($0.8193$ \geneval), leading on nearly every alignment, compositional, and no-reference quality metric. Narrowing the cap to $U(.02,.20)$ or widening it to $U(.02,.70)$ both erode these ($82.720$/$82.426$ \dpg, $0.8095$/$0.8113$ \geneval); raising the lower edge with $U(.15,.70)$ erodes them further ($82.038$ \dpg, $0.7924$ \geneval); and pushing supervision fully into the low-noise regime $U(.60,.90)$ collapses them ($80.330$ \dpg, $0.7341$ \geneval; color-attribute binding $0.735\!\to\!0.538$). This pattern supports selecting the high-noise band for alignment and compositionality.

\begin{table}[h]
\centering
\caption{Noise selection: matching-band position, width, and per-stage nesting. Four-step DINO-Adv students ($20$k, $\lambda_{\mathrm{GAN}}{=}.01$, AF-Loss disabled) differ only in the DMD matching range; $\uparrow$/$\downarrow$ give direction and bold is best per row. The high-noise band $U(.02,.35)$ leads on nearly every alignment metric.}
\label{tab:bandpos}
\small
\setlength{\tabcolsep}{7pt}
\resizebox{\textwidth}{!}{%
\begin{tabular}{@{}lrrrrrrr@{}}
\toprule
DMD matching range & $U(.02,.20)$ & $U(.02,.35)$ & $U(.02,.70)$ & $U(.15,.70)$ & $U(.60,.90)$ & $U(.02,.98)$ & $U(.02,\{.25,.50,.75,.98\})$ \\
 & \emph{narrow} & \emph{best} & \emph{one-sided} & \emph{wide} & \emph{low-noise} & \emph{full range} & \emph{per-stage} \\
\midrule
\geneval$\uparrow$ & 0.8095 & \textbf{0.8193} & 0.8113 & 0.7924 & 0.7341 & 0.8155 & 0.8164 \\
\dpg$\uparrow$ & 82.720 & \textbf{82.959} & 82.426 & 82.038 & 80.330 & 82.550 & 82.018 \\
CLIP$\uparrow$ & 0.3201 & \textbf{0.3225} & 0.3184 & 0.3213 & 0.3194 & 0.3214 & 0.3211 \\
IS$\uparrow$ & 38.79 & \textbf{40.21} & 39.78 & 39.30 & 36.21 & 39.38 & 38.71 \\
Recall$\uparrow$ & 0.4182 & \textbf{0.4318} & 0.4198 & 0.4062 & 0.3773 & 0.4297 & 0.4027 \\
NIQE$\downarrow$ & 3.257 & \textbf{3.034} & 3.296 & 3.424 & 3.636 & 3.337 & 3.596 \\
\bottomrule
\end{tabular}}
\end{table}

\paragraph{Band width: the cap sits at $.35$.} Fixing the lower edge at the numerical floor $.02$ and sweeping only the \emph{upper} edge $b$ of a uniform band $U(.02,b)$ traces out the cap directly (Table~\ref{tab:bandpos}). DPG rises from $82.720$ at $b\!=\!.20$ to a peak of $82.959$ at $b\!=\!.35$, then declines as the band admits more low-noise, texture-shortcut mass ($82.426$ at $b\!=\!.70$, $82.550$ at $b\!=\!.98$, i.e.\ nearly the full range). The best tested cap coincides with the $t_m\!=\!.35$ transition suggested by the representation diagnostic (Section~\ref{sec:dmd}), and widening past it degrades prompt following.

\paragraph{Single-step supervision.} Representation supervision becomes decisive at one step. The full anchor recipe scores $82.963$ \dpg\ and $0.7914$ \geneval, whereas the exact one-step no-GAN retrain reaches only $78.980$ \dpg\ and $0.7590$ \geneval. The frozen-DINOv2 signals thus recover most of the alignment that is lost when the trajectory collapses to a single step, though the balanced four-step model still leads on most reported metrics.

\section{Anchor-Field ablations}
\label{app:affeat}

\paragraph{AF-Loss implementation details.}\label{app:afimpl}
Algorithm~\ref{alg:af} gives the exact field used in our runs; the text below states its gradient-affecting choices. \textbf{Reference clouds:} the current all-gathered real/generated features are prepended to the rolling banks and truncated to $N_b{=}256$, so the pool spans the current and most recent batches; the banks are updated \emph{after} the generator step. \textbf{Scale $\bar d$:} a single detached scalar, the mean of the full query$\times$target $L_2$ distance matrix. \textbf{Symmetrized kernel / self-mask:} at each scale the affinity is the elementwise geometric mean of a softmax over targets and a softmax over queries of $-d(z_i,u)/(\bar d R)$ (doubly normalized); each $z_i$ is masked from its own copy among the negatives, the attraction and repulsion coefficients are cross-coupled, and each scale's displacement is RMS-normalized (the RMS is taken over the entire query$\times$feature matrix, per rank) and then summed over $\mathcal R{=}\{.02,.05,.2\}$. \textbf{Reduction:} $\mathcal L_{\mathrm{AF}}=\frac1B\sum_{i=1}^{B}\lVert z_i-\mathrm{sg}(z_i+V(z_i))\rVert_2^2$, the batch mean of the per-sample squared $L_2$. T he code matches Algorithm~\ref{alg:af}.

\begin{algorithm}[t]\small
\caption{AF-Loss: one generator step (auxiliary to the DMD/GAN updates)}
\label{alg:af}
\begin{algorithmic}[1]
\Require generator $G_\theta$; frozen encoder $\phi$; rolling banks $B_r,B_f$; radii $\mathcal R{=}\{.02,.05,.2\}$; pool $N_b{=}256$; weight $\lambda_{\mathrm{AF}}$
\State $\hat x\gets G_\theta(\epsilon,c)$;\ \ $z\gets\phi(\hat x)$ (query, keeps grad);\ \ $z_r\gets\phi(x)$ \Comment{$x$: real image}
\State all-gather (no grad): $\mathrm{cur}_r\gets z_r$,\ \ $\mathrm{cur}_f\gets z$ \Comment{local rows placed first}
\State $P\gets[\mathrm{cur}_r;B_r]_{1:N_b}$,\ \ $Q\gets[\mathrm{cur}_f;B_f]_{1:N_b}$ \Comment{current batch is included in the field}
\State $T\gets[\,Q\,;\,P\,]$ (negatives then positives);\ \ $D_{iu}\gets\lVert z_i-T_u\rVert_2$
\State $\bar d\gets\operatorname{mean}(D)$;\ \ $\tilde D\gets D/\bar d$ \Comment{$\bar d$ detached}
\State self-mask: $\tilde D_{ii}\mathrel{+}{=}c$ (large) on the $|Q|$ negative columns \Comment{$z_i$ vs its own copy}
\State $V\gets 0$
\For{$R\in\mathcal R$}
  \State $A\gets\sqrt{\operatorname{softmax}_{\mathrm{targets}}(-\tilde D/R)\odot\operatorname{softmax}_{\mathrm{queries}}(-\tilde D/R)}$ \Comment{symmetrized affinity}
  \State $A^-\!\gets\!A_{:,\le|Q|}$,\ $A^+\!\gets\!A_{:,>|Q|}$;\ \ $C\gets\big[\,-A^-\!\odot\!\textstyle\sum_u A^+\ \big|\ A^+\!\odot\!\textstyle\sum_u A^-\,\big]$ \Comment{cross-coupled}
  \State $F\gets C\,T-(\textstyle\sum_u C)\odot z$;\ \ $V\mathrel{+}{=}\,F/\operatorname{RMS}(F)$ \Comment{per-$R$ force norm}
\EndFor
\State $\mathcal L_{\mathrm{AF}}\gets\tfrac1B\sum_i\lVert z_i-\operatorname{sg}(z_i+V_i)\rVert_2^2$;\ \ add $\lambda_{\mathrm{AF}}\nabla_\theta\mathcal L_{\mathrm{AF}}$ to DMD/GAN grads; update $G_\theta$ \Comment{grad only $z\!\to\!\hat x\!\to\!G_\theta$}
\State $B_r\gets[\mathrm{cur}_r;B_r]_{1:N_b}$,\ \ $B_f\gets[\mathrm{cur}_f;B_f]_{1:N_b}$ \Comment{detached FIFO update, for next step}
\end{algorithmic}
\end{algorithm}

\textbf{Figure~\ref{fig:ofat} protocol.} Each panel is a single-factor sweep around a fixed four-step, 20k-step anchor with the high-noise band $U(.02,.35)$: the AF-Loss weight/bank/radii panels keep DINO-Adv off and AF-Loss on (bank $256$, radii $\{.02,.05,.2\}$, $\lambda_{\mathrm{AF}}{=}.05$ except the swept factor), while the DINO-Adv-weight panel keeps AF-Loss off. Each point is a single 20k run rather than a three-seed mean, so it selects defaults but is not directly comparable to the full-model numbers in Tables~\ref{tab:main}--\ref{tab:ablation}.

\paragraph{Why this field rather than another feature-space objective?}
To test whether the gain comes simply from applying a frozen-DINOv2 penalty, and whether any distribution-matching objective in the same space would do, we compare AF-Loss against two alternatives on the identical normalized last-layer CLS features, encoder, four-step protocol, and loss weight ($.05$): (i) direct perceptual matching (regression to real CLS features), and (ii) a matched multi-scale \emph{DINOv2-CLS MMD} that pulls the generated feature distribution toward the same $256$-feature real bank with the same kernel bandwidths, differing from AF-Loss only in that it minimizes a batch-level distribution distance rather than injecting a sample-dependent attract--repel displacement. All rows use the AF-Loss-only ablation setting in Table~\ref{tab:ablation}: DM-Band and DINO-Adv are disabled, and only the feature-space objective is changed. All three add no learnable parameters. Table~\ref{tab:afperceptual} shows AF-Loss is best on five of the seven metrics; direct regression attains the highest \geneval\ and MMD the highest recall. Against the stronger MMD baseline it raises DPG by $1.317$ ($83.194$ vs $81.877$), IS by $3.69$, VQA by $0.0130$, and CLIP by $0.0040$, approximately matches it on \geneval\ ($0.8090$ vs $0.8077$), and lowers NIQE by $0.859$; MMD attains slightly higher recall ($0.3330$ vs $0.3161$), indicating marginally broader coverage at the cost of alignment and quality. Direct regression gives higher \geneval\ but is worse on the other six metrics. The gains therefore do not arise from introducing DINOv2 features, nor from distribution matching per se; they depend on the specific sample-dependent attract--repel field.
\begin{table}[h]
\centering\scriptsize
\setlength{\tabcolsep}{2.5pt}
\resizebox{\columnwidth}{!}{%
\begin{tabular}{@{}lrrrrrrr@{}}
\toprule
Objective & \geneval$\uparrow$ & DPG$\uparrow$ & VQA$\uparrow$ & IS$\uparrow$ & CLIP$\uparrow$ & Rec.$\uparrow$ & NIQE$\downarrow$ \\
\midrule
\textbf{AF-Loss (CLS field, ours)} & 0.8090 & \textbf{83.194} & \textbf{0.7079} & \textbf{39.46} & \textbf{0.3246} & 0.3161 & \textbf{3.574} \\
Direct DINOv2-CLS perceptual & \textbf{0.8156} & 82.077 & 0.7028 & 37.85 & 0.3217 & 0.3146 & 4.675 \\
DINOv2-CLS MMD (distribution matching) & 0.8077 & 81.877 & 0.6949 & 35.77 & 0.3206 & \textbf{0.3330} & 4.433 \\
\bottomrule
\end{tabular}}
\caption{AF-Loss versus two alternative feature-space objectives on the same normalized last-layer CLS features, encoder, four-step protocol, and loss weight ($.05$): direct DINOv2-CLS perceptual matching, and a matched multi-scale DINOv2-CLS MMD (distribution matching) against the same $256$-feature real bank. Only the feature-space objective changes; \textbf{bold} is best per column.}
\label{tab:afperceptual}
\end{table}

\paragraph{Which feature.} Table~\ref{tab:affeat} ablates which frozen-DINOv2 feature the Anchor Field acts on, holding the field form and weight fixed at $\lambda_{\mathrm{AF}}\!=\!.05$. The normalized final-layer CLS feature, our default, gives the best DPG.
\begin{table}[h]
\centering\small
\setlength{\tabcolsep}{6pt}
\begin{tabular}{@{}p{0.62\linewidth}c@{}}
\toprule
AF feature & DPG\,$\uparrow$ \\
\midrule
DINOv2 $\{2,5,8,11\}$ spatial patch features (per-layer mean-pool, concat) & 81.970 \\
DINOv2 $\{2,5\}$ spatial patch-feature mean+std & 81.430 \\
DINOv2 $\{2,5\}$ spatial patch-feature quantile & 81.760 \\
\textbf{DINOv2 normalized final-layer CLS feature (ours)} & \textbf{83.194} \\
\bottomrule
\end{tabular}
\caption{Anchor-Field feature ablation. All variants share the same field form at $\lambda_{\mathrm{AF}}\!=\!.05$; the normalized final-layer CLS feature is our default.}
\label{tab:affeat}
\end{table}

\paragraph{Attract vs.\ repel.} Table~\ref{tab:afterm} isolates the two halves of the field, keeping the feature and weight fixed: attracting the student toward the real bank $B_r$, repelling it from the fake bank $B_f$, and both together (our default). Attraction toward the real bank supplies most of the gain, repulsion from the fake bank alone is markedly weaker, and the full attract--repel field is best; we therefore keep both terms as our default.
\begin{table}[h]
\centering\small
\setlength{\tabcolsep}{8pt}
\begin{tabular}{@{}lc@{}}
\toprule
Anchor-Field term & DPG\,$\uparrow$ \\
\midrule
Attract only (real bank $B_r$) & 82.910 \\
Repel only (fake bank $B_f$) & 80.170 \\
\textbf{Attract $+$ repel (ours)} & 83.194 \\
\bottomrule
\end{tabular}
\caption{Anchor-Field attract/repel ablation, at the default feature and $\lambda_{\mathrm{AF}}\!=\!.05$.}
\label{tab:afterm}
\end{table}

\section{Additional Qualitative Comparisons}
\label{app:qual}
\begin{figure*}[t]
\centering
\setlength{\tabcolsep}{1pt}
\begin{tabular}{@{}p{0.487\linewidth}@{\hspace{4pt}}p{0.487\linewidth}@{}}
\qmh & \qmh \\[2pt]
\qpl{An intricate oil painting that captures two rabbits standing upright in a pose reminiscent of the iconic American Gothic portrait, in early 20th-century rural clothing\,\ldots}
& \qpl{A rider atop a chestnut horse in the middle of a spacious pasture enclosed by a wooden fence, dotted with patches of green grass\,\ldots} \\
\qi{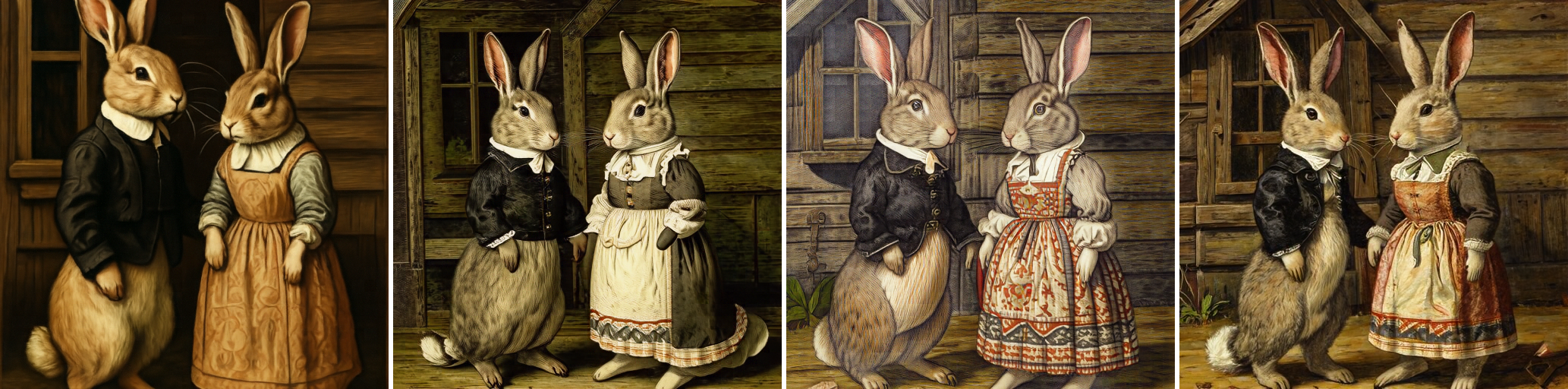} & \qi{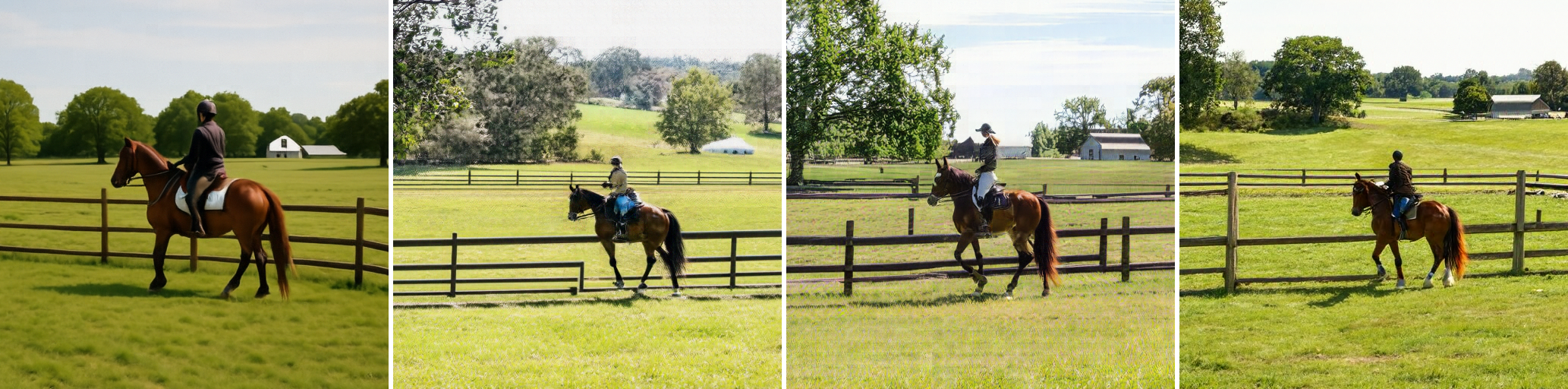} \\[5pt]
\qpl{A frisky golden retriever with a shiny, shaggy coat stands next to a life-sized penguin statue in the midst of a bustling public park\,\ldots}
& \qpl{A vibrant yellow rabbit, its fur almost glowing with cheerfulness, bounds energetically across a sprawling meadow, its sizeable red-framed glasses slipping comically\,\ldots} \\
\qi{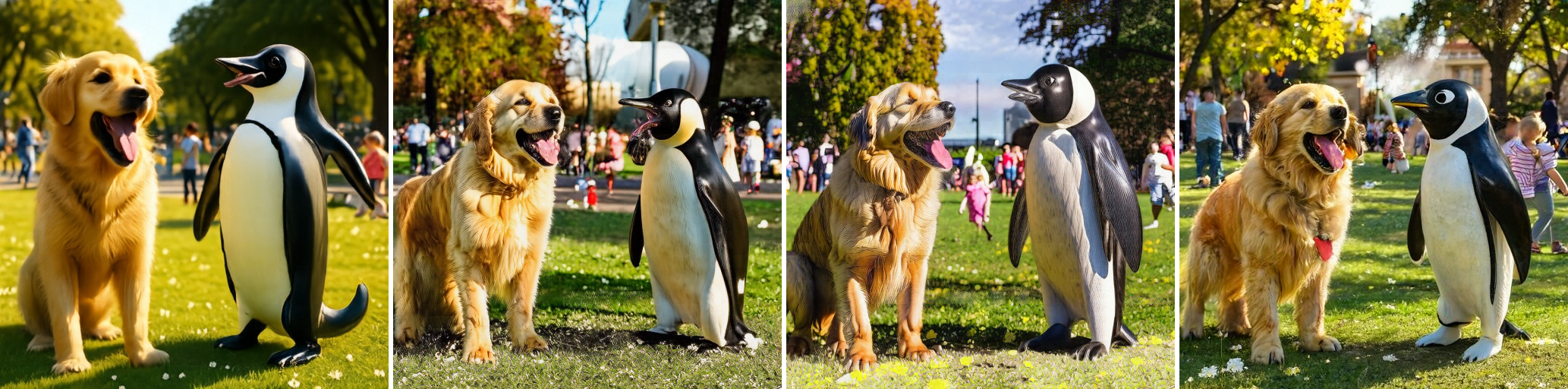} & \qi{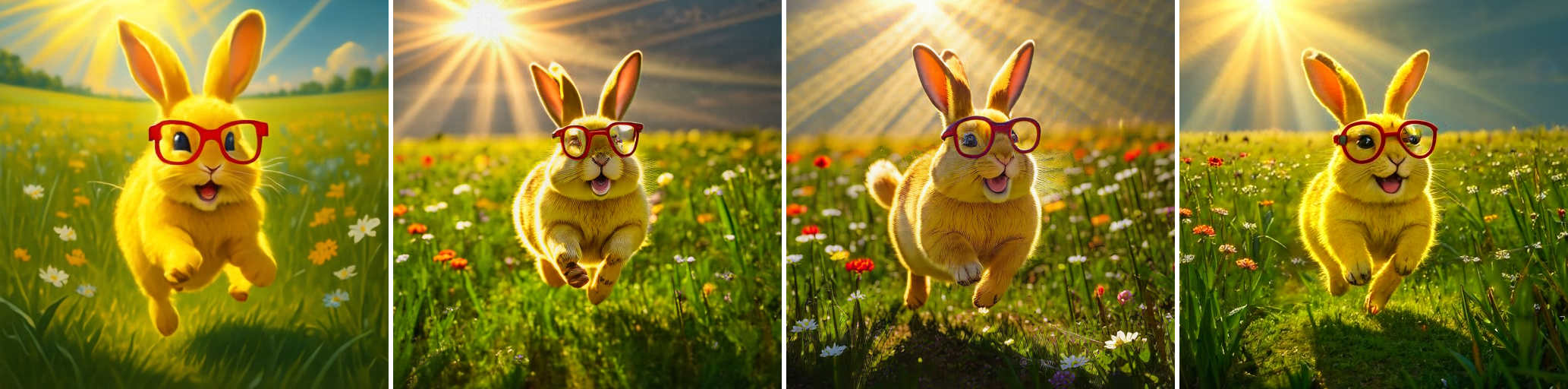} \\[7pt]
\qmh & \qmh \\[2pt]
\qps{A polar bear walking over rocks in its enclosure.} & \qps{A teddy bear wearing a green robe.} \\
\qi{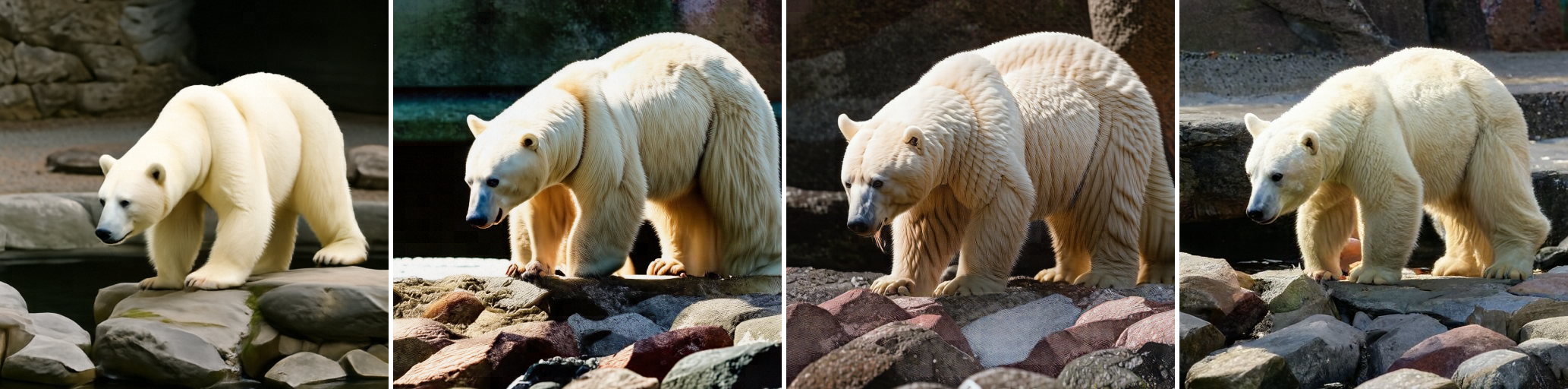} & \qi{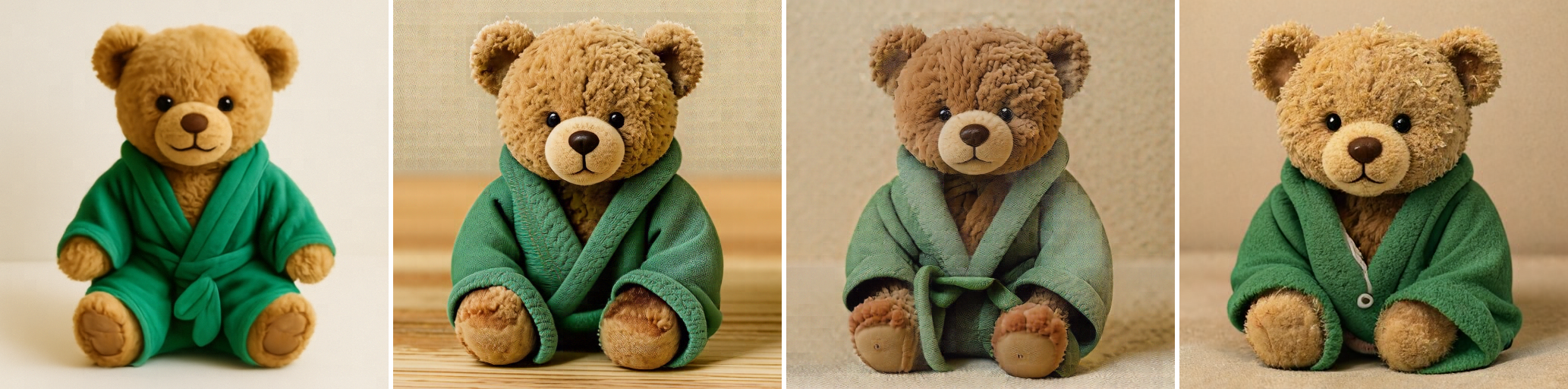} \\[5pt]
\qps{Two sheep standing next to each other in the snow.} & \qps{There is a dog that is walking on the beach at sunset.} \\
\qi{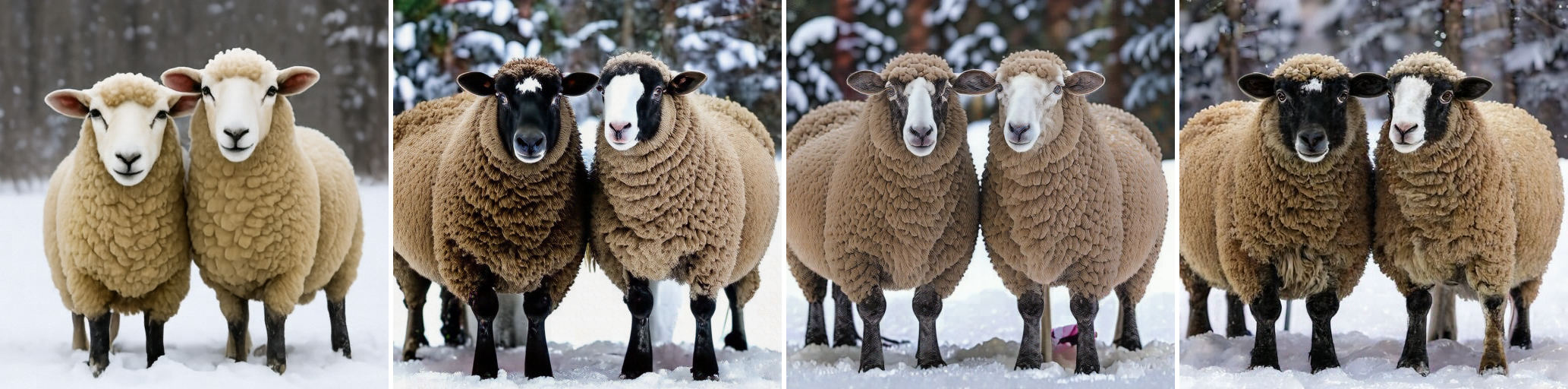} & \qi{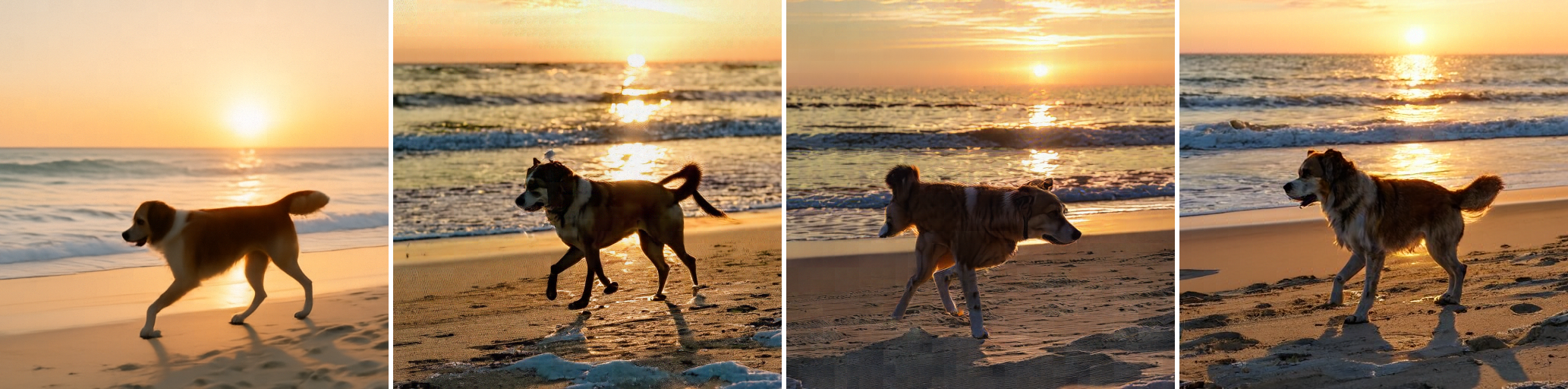} \\
\end{tabular}
\caption{Additional four-step qualitative comparisons under identical prompts (columns: 25-step teacher, DMD2, Decoupled, and \method{}). Top block: long \dpg\ prompts (abbreviated with ``\ldots''); bottom block: short COCO captions.}
\label{fig:qual_supp}
\end{figure*}

Figure~\ref{fig:qual_supp} shows eight further four-step comparisons under matched prompts, complementing Figure~\ref{fig:qual}. As in the main text, \method{} keeps object count, anatomy, and composition correct (two rabbits, a single rider on one horse, the dog beside the penguin, the two sheep) while the critic-coupled baselines more often duplicate or distort subjects.

\section{Limitations}
Two limitations remain. First, DINO-Adv and AF-Loss inherit the inductive biases of the frozen DINOv2 representation. Although its multi-layer patch features and final-layer CLS feature provide complementary local and semantic guidance, they may not fully capture text-conditioned spatial relations or domain-specific visual attributes. Second, our evaluation emphasizes established automated benchmarks and qualitative comparisons; broader human-preference, fairness, and safety evaluations remain future work.

\end{document}